\documentclass[10pt,journal,compsoc]{IEEEtran}
\ifCLASSOPTIONcompsoc
  \usepackage[nocompress]{cite}
\else
  \usepackage{cite}
\fi
\usepackage{soul}
\usepackage{url}
\usepackage[hidelinks]{hyperref}
\usepackage[utf8]{inputenc}
\usepackage{graphicx}
\usepackage{amsmath}
\usepackage{tabu}
\usepackage{amsthm}
\usepackage{booktabs}
\usepackage{algorithm}
\usepackage{algorithmic}
\usepackage{subfigure}
\usepackage{xspace}
\usepackage[dvipsnames]{xcolor}
\usepackage{multirow}
\usepackage{enumitem}
\usepackage{bm}

\usepackage{xcolor,colortbl}
\usepackage{caption}
\usepackage{amsmath,amsfonts,amssymb}
\usepackage{listings}
\usepackage{makecell}
\usepackage{savesym}
\savesymbol{checkmark}
\usepackage{multirow}
\usepackage{dingbat}
\usepackage{arydshln}
\usepackage{hyperref}
\usepackage{pifont}
\usepackage[figuresright]{rotating}
\usepackage{hyperref}
\usepackage[edges]{forest}
\hypersetup{
    urlcolor=blue
}
\usepackage{arydshln}
\usepackage{rotating}
\usepackage{tablefootnote}

\newcommand{\cmark}{\ding{51}}%
\newcommand{\xmark}{\ding{55}}%

\newcommand{\etal}{\emph{et al.}\xspace}
\newcommand{\ignore}[1]{}

\newcommand{\changed}[1]{\textcolor{black}{#1}}

\newcommand{\ysadd}[1]{\textcolor{black}{#1}}

\renewcommand{\figurename}{FIGURE}

\ifCLASSINFOpdf
\else
\fi
\begin{document}
%
\title{Survey of Natural Language Processing for Education: Taxonomy, Systematic Review, and Future Trends}
%
%
%
%

\author{Yunshi~Lan*, Xinyuan~Li, Hanyue~Du, Xuesong~Lu, Ming~Gao, Weining~Qian and~Aoying Zhou,
\IEEEcompsocitemizethanks{
\IEEEcompsocthanksitem Yunshi Lan, Xinyuan~Li, Hanyue~Du, Xuesong~Lu, Ming~Gao, Weining~Qian and~Aoying Zhou are with School of Data Science and Engineering, East China Normal University.
}

\thanks{\\
Manuscript revised xxx.\\
This work was supported in part by National Natural Science Foundation of China under Grants 62537001, in part by Guizhou Provincial Program on Commercialization of Scientific and Technological Achievements  (Qiankehezhongyindi [2025] No. 006), and in part by
the Chenguang Program of Shanghai Education Development Foundation and Shanghai Municipal Education Commission under Grant 24CGA26.\\
(Corresponding author: Yunshi Lan.)}}

%
%

\markboth{Journal of \LaTeX\ Class Files,~Vol.~14, No.~8, August~2015}%
{Shell \MakeLowercase{\textit{et al.}}: Bare Demo of IEEEtran.cls for Computer Society Journals}

\IEEEtitleabstractindextext{%
\begin{abstract}
Natural Language Processing (NLP) aims to analyze text or speech via techniques in the computer science field. It serves applications in the domains of healthcare, commerce, education, and so on. Particularly, NLP has been widely applied to the education domain and its applications have enormous potential to help teaching and learning. In this survey, we review recent advances in NLP with a focus on solving problems relevant to the education domain. In detail, we begin with introducing the related background and the real-world scenarios in education to which NLP techniques could contribute. Then, we present a taxonomy of NLP in the education domain and highlight typical NLP applications including question answering, question construction, automated assessment, and error correction.  Next, we illustrate the task definition, challenges, and corresponding cutting-edge techniques based on the above taxonomy. In particular, LLM-involved methods are included for discussion due to the wide usage of LLMs in diverse NLP applications. After that, we showcase some off-the-shelf demonstrations in this domain, which are designed for educators or researchers. At last, we conclude with five promising directions for future research, including generalization over subjects and languages, deployed LLM-based systems for education, adaptive learning for teaching and learning, interpretability for education, and ethical consideration of NLP techniques. We organize all relevant datasets and papers in the open-available Github Link for better review~\url{https://github.com/LiXinyuan1015/NLP-for-Education}.
\end{abstract}

}

\maketitle

\IEEEdisplaynontitleabstractindextext

%
\IEEEpeerreviewmaketitle

\section{Introduction}\label{sec1}






Natural Language Processing (NLP) refers to the branch of computer science and artificial intelligence, that empowers computers the ability to understand text and spoken words the way human beings can.
It serves real-world scenarios.
For example, NLP can help in \textit{healthcare} applications.
A wide range of NLP techniques are being utilized to anticipate the treatment process of the patients.
Question answering~\cite{lu:nips2022,chen:arxiv2023} and dialogue generation~\cite{wu:acl2020} help conduct medical diagnosis, administrative assistants for appointments, and billing.
Similarly, NLP can help in \textit{commerce} applications.
There is a growing trend of incorporating NLP techniques into analytics and customer management platforms to help learn more about the relationship between the consumers and items via sentiment analysis~\cite{yu:emnlp2016}.

In the education domain, the widespread collaboration of new technologies into the \textit{education} domain revolutionizes education in the future.
NLP techniques can help education in the following ways: 
\begin{itemize}
    \item NLP technologies provide automatic key information extraction from unstructured text which facilitates highlighting the key points when teachers and students read passages, tutorials, and textbooks.
    \item NLP technologies offer intelligent tutoring and recommendations.
    When students are learning a concept, a system can help recommend relevant concepts, practices, and reading materials such that the students can find proper reading materials as references, which enables active learning and self-inspection for students.
    \item NLP technologies help automatic problem-solving.
    For example, in math or science courses, students may have extensive questions about exercises after the classes.
    To avoid repeated and overwhelming questions for teachers, NLP technologies can conduct reasoning and answer the questions automatically.
    \item NLP technologies facilitate automatic smart content generation when teachers have trouble in preparing the teaching materials from a rush of information on the Internet or textbooks.
    Additionally, given a passage in the textbook as the context and a concept as the correct answer, NLP techniques can quickly construct a multiple-choice question associated with the test question and the distractors.
    This could facilitate the teachers to make quick quizzes for classes.
    \item NLP techniques enable personalized instruction.
    Students may have different background and it is vital to track the learning trajectory and provide personalized instruction to the students.
    Various NLP models trained on different annotated data collected from different scenarios could contribute to personalized education.
    \item NLP technologies enable AI-supported assessment and correction.
    For example, it is time-consuming for teachers to grade the essays and codes from students in linguistics or coding classes, by analyzing the textual contents of the assignments, NLP applications can figure out the errors in the assignment and return the corresponding correction to the students.
\end{itemize}
The above techniques allow for more flexible and effective learning for students, thus freeing up valuable time for teachers.

We notice there are a few surveys related to NLP for education.
However, a line of surveys focuses on a certain task.
Specifically, Lan~\etal~\cite{lan_tkde2023} summarize the general advancement for knowledge base question answering tasks but the domain is not restricted to education.
Bryant~\etal~\cite{Bryant_Yuan} summarize the cutting-edge techniques of grammatical error correction, which is a typical task of educational NLP.
Zhang~\etal~\cite{zhang:tpami2019} carefully review the technical progress of math word problem solving tasks.
Messer~\etal~\cite{messer:tce2023} provide a systematic review of the task of code programming.
However, the above surveys focus solely on a specific task and do not identify the significant distinction of these tasks applied in the education domain.
\changed{Recently, we found there are some surveys conducting a specific discussion on large language models for education~\cite{li:arxiv2023,wang:arxiv2024}, which are advanced but cover only partial NLP techniques}.

To provide a comprehensive overview of NLP for education from a technical point of view: (1) \textbf{New taxonomy}: we present a novel taxonomy for educational NLP and identify representative NLP tasks that can be applied for learning and comprehension as well as writing and assessment in education scenarios, and then discuss them in detail;
(2) \textbf{Comprehensive review}: we collect recent papers published in well-recognized computer science venues and review their methods for solving the challenges of corresponding tasks.
We further include LLM-based methods due to their wide usage in NLP applications;
(3) \textbf{Rich resources}: we discuss the publicly-available datasets and demonstrations with their various attributes, which makes it easy for readers to get started for their research.


For the rest of the manuscript, we first present the taxonomy of NLP in the education domain, which highlights four representative tasks and eight fine-grained sub-tasks in total.
For each task, we review the datasets and techniques for the identified tasks, pointing out the major challenges and corresponding solutions.
Then, we summarize the available demonstrations.
At last, we conclude the survey with future directions.

\begin{figure}[t!] 
	\centering
	\includegraphics[width=.47\textwidth]{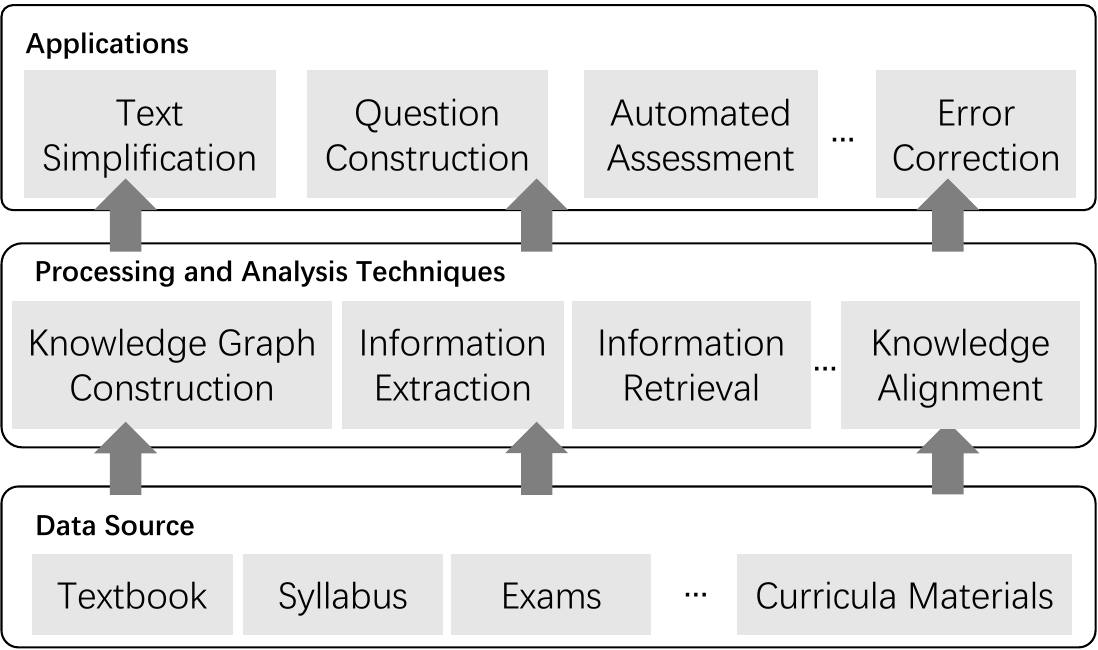}
	\caption{An overview of NLP for education.
 The left section displays the systematic frame of relevant NLP techniques, including data sources, processing and analysis techniques, and applications.
 The right section demonstrates the outcomes of diverse techniques.
    }
	\label{fig:significance}
\end{figure}

\begin{figure*}
\centering
	\footnotesize
	\begin{forest}
		for tree={
			forked edges,
			grow'=0,
			draw,
			rounded corners,
			node options={align=left},
			calign=edge midpoint,
		},
	    [NLP Applications for Education, text width=2.0cm, for tree={fill=white!20}
			[Textbook Question Answering, text width=1.7cm, for tree={fill=white!20}
				[The subject-specific QA data is limited for training, text width=3.8cm, for tree={fill=white!20}
				    [
				    Fine-tune the contextual knowledge as language modeling~\cite{ram:arxiv2021,xu:arxiv2022} and augment data from different subjects~\cite{lee:mrqa2019}.,
				    text width=7.5cm, node options={align=left}
				    ]
				]
				[The contexts may be presented in diverse formats, text width=3.8cm, for tree={fill=white!20}
				    [ 
				    Incorporate the pre-processing steps of diagrams/tables/graphs as off-the-shelf toolkits~\cite{poco:cgf2017,kafle:cvpr2018}.,
				    text width=7.5cm, node options={align=left}
				    ]
				]
				[Zero-shot issues frequently occur, text width=3.8cm, for tree={fill=white!20}
				    [Collaborate with LLMs as the agent in TQA tasks and integrate the image-to-caption modules and pre-processing toolkits~\cite{wang_mtqa:arxiv2023,liu:arxiv2023,wang:aaai2024}.,
				    text width=7.5cm, node options={align=left}
				    ]
				]
			]
			[Math Word Problem Solving, text width=1.7cm, for tree={fill=white!20}
				[Some MWPs are complex in logic, text width=3.8cm, for tree={fill=white!20}
                    [Formulate the math expression as an abstract tree and re-organize the textual contexts with structures~\cite{wang:emnlp2018,chiang:naacl2019,liu:emnlp2019,xie:ijcai2019,zhang:acl2020,lin:aaai2021}.,
				    text width=7.5cm, node options={align=left}
				    ]
				]
				[There is a lack of annotated expressions in the MWPs, text width=3.8cm, for tree={fill=white!20}
				    [ 
				    Learn the MWPs solving with weak supervision~\cite{wang:emnlp2018,hong:aaai2021}; Distill the knowledge from a large pre-trained generic model~\cite{zhang:ijcai2020}.,
				    text width=7.5cm, node options={align=left}
				    ]
				]
				[Zero/few-shot complex MWPs in real-world scenarios, text width=3.8cm, for tree={fill=white!20}
				    [ 
				    Force LLMs to think step by step via CoT~\cite{Wei:arxiv2023,Wang:arxiv2023,Zhang:arxiv2022, Zhou:arxiv2022,Zheng:arxiv2023,lu:arxiv2023}; Treat LLMs as agent to call toolkits for processing contexts in multi-modalities~\cite{chen:tmlr2023}.,
				    text width=7.5cm, node options={align=left}
				    ]
				]
            ]
			[Question Generation, text width=1.7cm, for tree={fill=white!20}
				[The generated questions ignore the difficulty stratification, text width=3.8cm, for tree={fill=white!20}
				    [
                    Setup a QA model as the discriminator~\cite{gao2018difficulty}; Propose a feature of difficulty levels for learning based on the number of inference steps/ item response theory/ Bloom's Taxonomy~\cite{cheng2021guiding,uto2023difficulty,forehand2005bloom}; Prompt LLMs to indicate difficulty levels~\cite{lee2024few,maity:fire2024}.,
				    text width=7.5cm, node options={align=left}
				    ]
				] 
				[The generated questions show weak relevance to the syllabus, text width=3.8cm, for tree={fill=white!20}
				    [
                    Setup a binary model as the discriminator to judge if it is related to certain keywords or concepts\cite{steuer2021not,hadifar2023diverse,xiao2023evaluating} which are various for subjects.,
				    text width=7.5cm, node options={align=left}
				    ]
				]
				[The generated questions should be customized to students, text width=3.8cm, for tree={fill=white!20}
				    [Incorporate a knowledge tracking model~\cite{srivastava2021question,wang2023difficulty} with pre-defined profiles of students.,
				    text width=7.5cm, node options={align=left}
				    ]
				]
			]
			[Distractor Generation, text width=1.7cm, for tree={fill=white!20}
				[The produced distractors are irrelevant to the questions, text width=3.8cm, for tree={fill=white!20}
				    [Employ dual attention and contrastive learning to ensure the consistency and high relevance to the contexts~\cite{bitew2023distractor, wang2023multi, yu2024enhancingdistractorgenerationmultiplechoice,gao2019generating,ding2024can}.
				    ., text width=7.5cm, node options={align=left}
				    ]
				]
				[The distractors lack strong distractive effectiveness, text width=3.8cm, for tree={fill=white!20}
				    [Propose methods to avoid excessive similarity between the generated distractors~\cite{alhazmi2024distractor,taslimipoor2024distractor,qu2024unsupervised}.
				    ., text width=7.5cm, node options={align=left}
				    ]
				]
			]
			[Automated Essay Scoring, text width=1.7cm, for tree={fill=white!20}
				[The singular evaluation of the essays lacks comprehensiveness, text width=3.8cm, for tree={fill=white!20}
				    [
			          Introduce more diverse evaluation metrics~\cite{persing:acl2013,persing:acl2014} and explainable prompts~\cite{chen:aied2024} considering specific objects and scenarios.,
				    text width=7.5cm, node options={align=left}
				    ]
				]
				[The sparse annotation of the essays in domain-specific prompts, text width=3.8cm, for tree={fill=white!20}
				    [ 
				    Decouple the domain-dependent and domain-independent features~\cite{jiang:acl2023,jiang:www2021,chen:acl2023}.,
				    text width=7.5cm, node options={align=left}
				    ]
				]
			]
			[Automated Code Scoring, text width=1.7cm, for tree={fill=white!20}
                [It is hard to identify good features and scoring algorithms, text width=3.8cm, for tree={fill=white!20}
				    [Collect comments/ number of lines/ number of operators to train a model to evaluate the quality of the code~\cite{lajis:icsca2018,glassman:tochi2015,talbot:sce2020,von:ie2018,queiros:aicse2012} with supervision
				    .,
				    text width=7.5cm, node options={align=left}
				    ]
				]
				[Implicitly encode the code snippet via distributed representation, text width=3.8cm, for tree={fill=white!20}
				    [Encode a code snippet and compare with the correct ones~\cite{wang:mlkdd2023,allamanis:fse2015,zhang:icse2019,alon:popl2019,hellendoorn:iclr2019,feng:arxiv2020,kanade:pmlr2020}
				    .,
				    text width=7.5cm, node options={align=left}
				    ]
				]
			]
			[Grammatical Error Correction, text width=1.7cm, for tree={fill=white!20}
				[GEC with LLMs encounters over-correction issue, text width=3.8cm, for tree={fill=white!20}
				    [Conduct ensemble of the edits produced via LLMs~\cite{li2024rethinking,wang2024lm,omelianchuk2024pillars}., 
				    text width=7.5cm, node options={align=left}
				    ]
				]
				[Distorted results of GEC systems in low resources, text width=3.8cm, for tree={fill=white!20}
				    [Construct multilingual data to handle mix-language text~\cite{chan2024grammatical}., 
				    text width=7.5cm, node options={align=left}
				    ]
				]
			]
			[Code Error Correction, text width=1.7cm, for tree={fill=white!20}
				[The structural information of code has been ignored, text width=3.8cm, for tree={fill=white!20}
                [Utilize tree structure~\cite{Guo_Ren_Lu_Feng_2020,Guo_Lu_Duan_Wang,Ahmad_Chakraborty_2021}/ AST data~\cite{wang-etal-2021-codet5} to model the structural characteristics of code.,
				    text width=7.5cm, node options={align=left}
				    ]
				]
				[CEC systems have resource overhead issue, text width=3.8cm, for tree={fill=white!20}
                [Incorporating grammar constraints of programming languages~\cite{Dong_Jiang_Liu_2022,anand2024comprehensivesurveyaidrivenadvancements}.,
				    text width=7.5cm, node options={align=left}
				    ]
				]
				[CEC systems encounter overfitting issue, text width=3.8cm, for tree={fill=white!20}
                [Evaluate on a wider range of execution scenarios~\cite{zhang2023surveylearningbasedautomatedprogram}; propose overfitting detection systems~\cite{nilizadeh2021exploring,Ye_2022},
				    text width=7.5cm, node options={align=left}
				    ]
				]
			]
		]
	\end{forest}
	\caption{\changed{The detailed techniques of NLP applications. The hierarchical structure is arranged with: NLP applications $\rightarrow$ Challenges $\rightarrow$ Solutions.}}
    \label{fig:taxonomy-tree}
\end{figure*}

\section{Taxonomy of NLP in Education Domain}
\label{sec:taxonomy}

\begin{figure}[t!] 
	\centering
	\includegraphics[width=.48\textwidth]{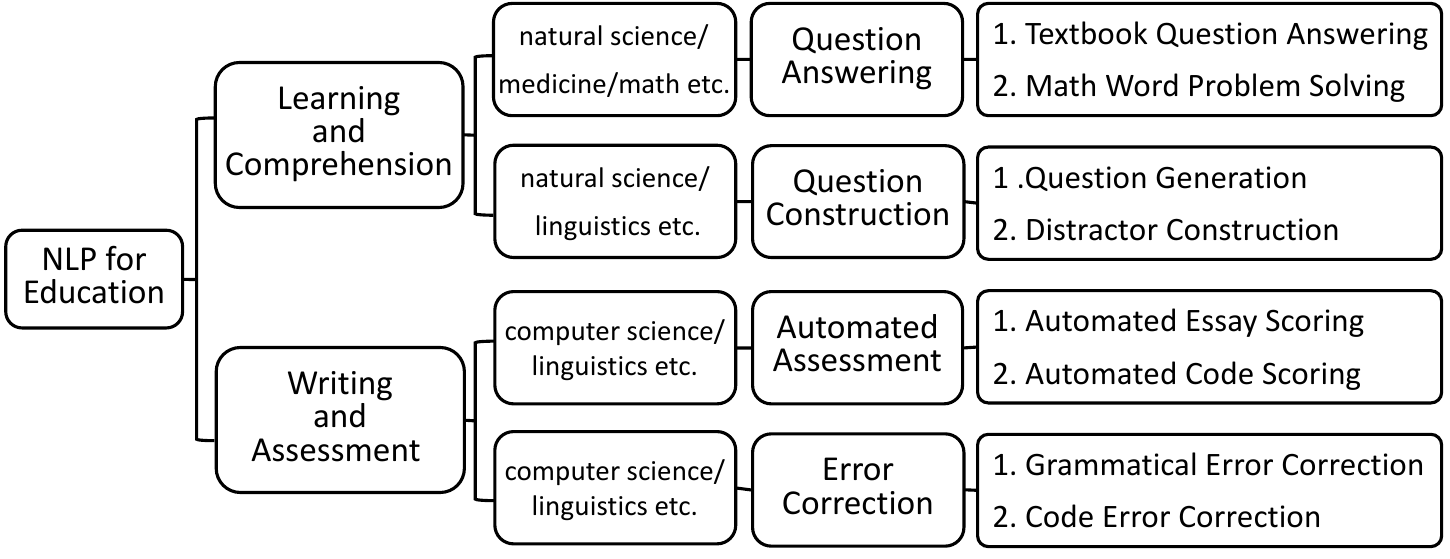}
	\caption{The taxonomy of NLP for Education.
 The NLP applications associated with the well-defined tasks and applicable subjects are displayed.
    }
	\label{fig:overview}
\end{figure}

Systems built on the basis of NLP applications and equipped with NLP techniques could perform as empowered AI tutoring, which enhances the teaching and learning experience of students and teachers.
With the systems, students are able to get access to private tutoring and extra support outside the classroom, which avoids increasing the workload of teachers.

We first demonstrate an overview of NLP for education in Figure~\ref{fig:significance}, where the systematic frame of relevant NLP techniques and their corresponding outcomes are displayed.
The data sources are significant as the fundamental supports for subsequent analyses and applications.
Generally, we could obtain the data from textbooks, syllabus, exams, and curricula materials, which may contain behavioral data of students, exercise questions, guidance signals of error correction, and so on.
From the perspective of teaching subjects, the data resources cover subjects like chemistry, math, linguistics, computer science, and so on, which could support applications in different subjects.
Given the raw data resources, we could conduct processing and analysis (e.g., knowledge graph construction, information extraction, information retrieval, knowledge alignment) to extract the key information from massive data and organize them in structured knowledge graphs (e.g., EduKG\footnote{\url{https://edukg.cn/team}}), databases (e.g.,  WordNet\footnote{\url{https://wordnet.princeton.edu/documentation}}) and lists, which could be applied to various downstream applications.
Upon the well-organized data, NLP tasks are applied (e.g., question answering, question construction, automated assessment, and error correction) to form high-level applications to serve teachers and students in real-world scenarios.
Grammarly\footnote{\url{https://app.grammarly.com}}, Gauthmath\footnote{\url{https://www.gauthmath.com}}, CodeSubmit\footnote{\url{https://codesubmit.io}}, AutoQG~\footnote{\url{https://autoqg.net}}, and EduChat\footnote{\url{https://www.educhat.top/}} are possible educational products and demonstrations developed based on these NLP applications.

This survey mainly focuses on the application level.
We highlight the most representative NLP applications with the taxonomy shown in Figure~\ref{fig:overview}.
Considering that NLP techniques play significant roles in different phases of education, on the one hand, NLP could serve as a teaching assistant for \textbf{learning and comprehension}.
When students learn the curriculum, the following NLP tasks could simplify the difficulties of learning. 

\begin{itemize}
\item \textbf{Question Answering (QA)}, as a typical application, could not only provide tutorials for students but also save time for overloaded teachers by providing feedback on questions.
Based on the different goals of question answering, textbook question answering and math word problem solving are presented as the two typical subcategories of question answering.
These two tasks both aim to generate answers from the given contexts automatically.
They can be widespread in the education of science subjects.
\item \textbf{Question Construction (QC)} is to generate questions as well as the distractors given the correct answers, which helps the teachers create a wide range of test questions using pre-existing learning resources.
Generally, a good test question may consist of a question and multiple distractors.
Therefore, question generation and distractor construction are two key sub-tasks to be reviewed.
\end{itemize}
On the other hand, NLP improves \textbf{writing and assessment} for students.
When given constructive criticism and writing prompts, students are likely to go deep into the topic.
However, teachers sometimes have difficulties responding thoughtfully to students due to time limitations.
But they will benefit from the following tasks.

\begin{itemize}
\item \textbf{Automated Assessment (AA)} aims to grade assignments for students automatically.
Language grading is a representative scenario of AA, which includes automated essay scoring for natural language learning and automated code scoring for program language learning.
For essay scoring, grammar, spelling, and sentence structure are key aspects that need to be considered.
For code scoring, the correctness of the code, style aspects, code quality are metrics to be viewed.
\item \textbf{Error Correction (EC)} for language are useful to improve language learning via explicating the correct edits.
Still, taking language learning as an example, grammatical error correction and code error correction are canonical sub-tasks for EC.
The methods are not simply required to provide error-free outputs based on the erroneous inputs, more detailed instructions should be provided to shed light on the reasons or knowledge points to the errors.
\end{itemize}

Please be aware that there are some other NLP tasks that are related to education.
For example, tasks like text rewriting~\cite{cripwell:aclfinding2023} and knowledge graph construction~\cite{wright:2022bioact} can also help teachers to prepare suitable teaching materials.
In this survey, we mainly focus on the highlighted tasks, which are widely explored NLP tasks in the education domain.
Our discussion is carried out based on the above taxonomy.

\section{Review of Techniques}

To help readers catch up with cutting-edge research on these highlighted tasks, in this section, we provide a comprehensive review of the evaluated resources and techniques for these tasks, showing emphasis on the development of techniques and solutions to the key challenges.
The review is arranged in a tree in Figure~\ref{fig:taxonomy-tree} for a quick lookup.

\subsection{Question Answering}

\begin{table*}[t!]
\centering
\resizebox{\textwidth}{!}{
\begin{tabular}{l c c c c | c c c}
        \toprule  
        Dataset & Science subject & Source & Context & \#Q  & Top 1 & Top 2 & Top 3 \\
        \midrule
        \multirow{2}{*}{TQA$\vartriangle$} & life/earth/ & grade 6-8  & \multirow{2}{*}{image/diagram} & \multirow{2}{*}{$26$K} & $84.2$~\cite{abdulrahman:arxiv2024} & $74.4$~\cite{abdulrahman:arxiv2024} & $72.1$~\cite{abdulrahman:arxiv2024}\\
        & physical science & science curricula & & &
        Llama-2 RAG~\cite{abdulrahman:arxiv2024} & SSCGN~\cite{wang:tcsvt2023} & ISAAQ~\cite{gomez:arxiv2020}  \\
        \hdashline
        \multirow{2}{*}{GeoSQA$\vartriangle$}  & \multirow{2}{*}{geography} & \multirow{2}{*}{high school} & \multirow{2}{*}{image/diagram} & \multirow{2}{*}{$13$K} & $26.2$\cite{huang:geosqa2019} & $25.9$\cite{huang:geosqa2019} & $25.2$\cite{huang:geosqa2019} \\
        & & & & & PMI~\cite{clark:aaai2016} & ESIM~\cite{chen:acl2017} & IR~\cite{clark:aaai2016} \\
        \hdashline
        \multirow{2}{*}{AI2D$\vartriangle$}  & \multirow{2}{*}{science} & \multirow{2}{*}{grade 1-6} & \multirow{2}{*}{image/diagram} & \multirow{2}{*}{$5$K} & $94.7$\cite{ye:arxiv2024} & $94.2$\cite{ye:arxiv2024} & $94.4$\cite{ye:arxiv2024}\\ 
        & & & & &  Claude 3.5 Sonnet & GPT-4o & Gemini-1.5-Pro  \\
        \hdashline
        \multirow{2}{*}{S{\tiny CIENCE}QA$\vartriangle$}  & natural/social/ & grade 1-12 & image/diagram/ & \multirow{2}{*}{$21$K} & $96.2$\cite{wang:aaai2024} & $94.9$\cite{tan:eccv2023} & $94.4$\cite{cha:cvpr2024} \\
        & language science & science curriculum & text &  &Multimodal-T-SciQ\cite{wang:aaai2024} & MC-CoT-F\cite{tan:eccv2023} & Honeybee\cite{cha:cvpr2024} \\
        \hdashline
        \multirow{2}{*}{MedQA$\vartriangle$} & \multirow{2}{*}{medicine} & professional medical & \multirow{2}{*}{text} &  \multirow{2}{*}{$40$K} & $86.5$\cite{jin:arxiv2020} & $81.4$\cite{jin:arxiv2020} & $67.6$\cite{jin:arxiv2020} \\
        & &  board exam & &  & MedPaLM-2 & GPT-4 & Flan-PaLM \\
        \hdashline
        \multirow{2}{*}{MedMCQA$\vartriangle$}  & \multirow{2}{*}{medicine} & \multirow{2}{*}{simulated exams} & \multirow{2}{*}{text} & \multirow{2}{*}{$200$K} & $63.0$\cite{lievin:icml2023} & $57.6$\cite{singhal:nature2022} & $56.5$\cite{singhal:nature2022} \\
        & & & & & VOD BioLinkBert & Med-Palm & Flan-PaLM \\
        \hdashline
        \multirow{2}{*}{TheoremQA$\vartriangle$}  & math/physics/ & \multirow{2}{*}{university exam} & \multirow{2}{*}{text} & \multirow{2}{*}{$800$} & $43.4$\footnotemark & $42.0$ & $34.9$ \\
        & EE/CS & & & & GPT-4 & LLaMA3-70B & Qwen1.5-110B \\
        \midrule 
        \multirow{2}{*}{Dolphin-18K$\dagger$}  & \multirow{2}{*}{math} & \multirow{2}{*}{grade school} & \multirow{2}{*}{text} & \multirow{2}{*}{$18$K} & $23.8/36.7/49.6$~\cite{qin:tkdd2024} & $15.8/31.2/40.7$~\cite{qin:tkdd2024} & $12.5/24.8/43.6$~\cite{qin:tkdd2024}\\
        & & & & & DMR~\cite{qin:tkdd2024} & SMR~\cite{qin:tkdd2024} & MaGNET~\cite{zhou:icnlg2019}\\
        \hdashline
        \multirow{2}{*}{DRAW-1K$\vartriangle$}  & \multirow{2}{*}{math} & \multirow{2}{*}{algebra} & \multirow{2}{*}{text} & \multirow{2}{*}{$1$K} & $63.5$~\cite{kim:acl2022} & $62.5$~\cite{kim:acl2022} & $59.5$~\cite{kim:acl2022} \\
        & & & & & EPT-XL~\cite{kim:acl2022} & GEO~\cite{ki:coling2020} & EPT-L~\cite{kim:acl2022} \\
        \hdashline
        \multirow{2}{*}{Math23K$\vartriangle$}  & \multirow{2}{*}{math} & \multirow{2}{*}{grade school} & text & \multirow{2}{*}{$23$K} & $94.3$~\cite{tan:arxiv2024} & $85.2$~\cite{zhang:emnlp2022} & $84.3$~\cite{shen:emnlp2021} \\
        & & & & & Teaching-GPT-4~\cite{tan:arxiv2024} & Multi-view~\cite{zhang:emnlp2022} & Generate\&Rank~\cite{shen:emnlp2021} \\
        \hdashline
        \multirow{2}{*}{MathQA$\vartriangle$}  & \multirow{2}{*}{math} & \multirow{2}{*}{grade school} & \multirow{2}{*}{text} & \multirow{2}{*}{$37$K} & $83.0$~\cite{zhang:nips2022} & $81.5$~\cite{wang:emnlp2018} & $80.6$~\cite{zhang:emnlp2022}\\
        & & & & & ELASTIC~\cite{zhang:nips2022} & Exp-Tree~\cite{wang:emnlp2018} & Multi-view~\cite{zhang:emnlp2022} \\
        \hdashline
        \multirow{2}{*}{ASDiv$\vartriangle$}  & \multirow{2}{*}{math} & \multirow{2}{*}{grade school} & \multirow{2}{*}{text} &  \multirow{2}{*}{$2$K} & $89.2$~\cite{hu:arxiv2024} & $40.4$~\cite{schick:nips2023} & $14.0$~\cite{schick:nips2023}\\
        & & & & & Self-Refine~\cite{hu:arxiv2024} & Toolformer~\cite{schick:nips2023} & GPT-3 \\
        \hdashline
        \multirow{2}{*}{GSM8K$\vartriangle$}  & \multirow{2}{*}{math} & \multirow{2}{*}{grade school} & \multirow{2}{*}{text} & \multirow{2}{*}{$8$K} & $97.1$~\cite{zhong:arxiv2024} & $94.6$~\cite{zhong:arxiv2024} & $94.3$~\cite{zhong:arxiv2024}\\
        & & & & & DUP-GPT-4~\cite{zhong:arxiv2024} & CoT-GPT-4~\cite{Wei:arxiv2023} & PS-GPT-4~\cite{wang:acl2023} \\
        \hdashline
        \multirow{2}{*}{IconQA$\vartriangle$}  & \multirow{2}{*}{math} & \multirow{2}{*}{K-12} & \multirow{2}{*}{text/image} & \multirow{2}{*}{$107$K} & $73.1$~\cite{kim:arxiv2024} & $55.9$~\cite{kim:arxiv2024} & $44.8$~\cite{kim:arxiv2024}\\
        & & & & & InstructBLIP~\cite{kim:arxiv2024} & LLaMA-2-Chat-7B & Vicuna-13B \\
        \hdashline
        \multirow{2}{*}{T{\tiny AB}MWP$\vartriangle$}  & \multirow{2}{*}{math} & \multirow{2}{*}{grade school}  & \multirow{2}{*}{text/table} & \multirow{2}{*}{$38$K} & $59.8$~\cite{zheng:acl2024} & $57.8$~\cite{zheng:acl2024} & $17.9$~\cite{zheng:acl2024} \\
        & & & & & Table-LLaVA 13B~\cite{zheng:acl2024} & Vicuna-1.5 7B & Llama2+Oracle \\
        \hdashline
        \bottomrule
    \end{tabular}
}
\caption{Comparison of TQA and MWP solving datasets.
The first section includes datasets of TQA.
The second section includes datasets of MWP solving.
``\#Q'' denotes the number of questions.
We select these datasets because they are constructed upon real-world textbook materials from different subjects.
The references of the reported results or the methods producing the results are shown at the right side or directly below the results, respectively.
\changed{All datasets are measured via the Answer Accuracy$\vartriangle$ except that Dolphin-18K is measured via BLEU-4/METEOR/ROUGE-L$\dagger$.
}
}
\label{tab:model&type}
\vspace{-0.5cm}
\end{table*}
\footnotetext{\url{https://huggingface.co/spaces/wenhu/Science-Leaderboard}}

\begin{figure}[th!] 
	\centering
	\includegraphics[width=0.5\textwidth]{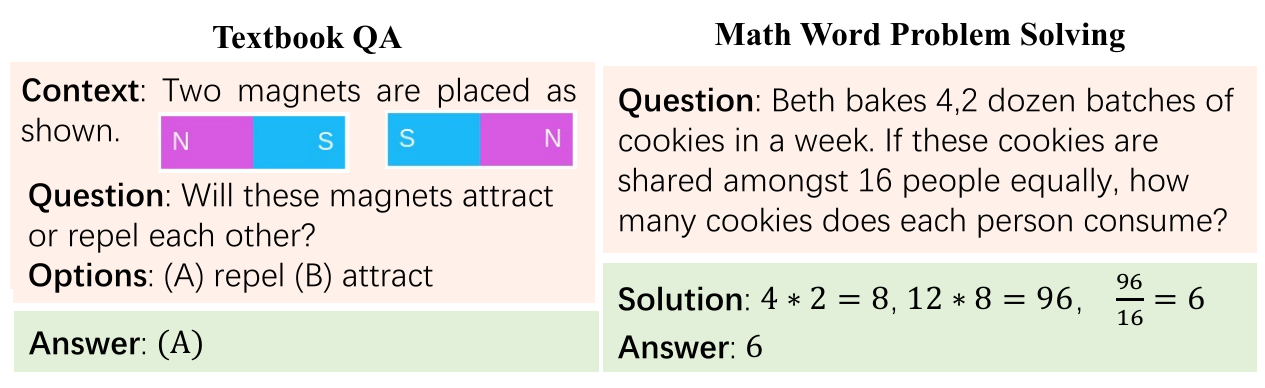} 
	\caption{The examples of question answering in education domain.
The red box denotes the inputs and the green box denotes the required outputs for the tasks.
 The displayed examples are extracted from S{\tiny CIENCE}QA and GSM8K, respectively.
    }
	\label{fig:eduqa}
\end{figure}

Even though QA has been explored as a popular NLP task for decades, educational QA requires a system to answer the domain-specific questions, related to natural science, social science, language science and so on~\cite{soares:imecs2021}.
These systems act as assistants to help answer difficult questions such that the students could learn from them and boost their performance.
Question answering in different subjects may be presented in different formats.
In Figure~\ref{fig:eduqa}, we display two typical examples of question answering in the education domain.
To answer the displayed questions, the systems need domain-specific knowledge like mathematics and science with the given contexts like text and images.
The output of the questions can be presented as an entity or a math expression.
In this section, we will present cutting-edge techniques for Textbook QA and MWP solving.

\subsubsection{Textbook Question Answering} 

\changed{Textbook Question Answering (TQA) is a task that requires a system to comprehensively understand the multi-modal information from the textbook curriculum, spreading across text documents, images, and diagrams.
We formulate the general loss function of TQA as follows for simplicity:
\begin{align*}
L_{\text{TQA}} = - \mathbb{E}_{(P, Q, A) \in \mathcal{D}} \log P_{\theta} (A| Q, P)
\end{align*}
where $P, Q, A$ are contexts, questions and answers from data collection $\mathcal{D}$, respectively.
$P$ may be presented in non-textual modalities.
The major challenge of TQA is to comprehend the multi-modal domain-specific contexts as well as the questions, and then identify the key information to the questions.
}

\noindent \textbf{Datasets.} 
Kembhavi~\etal~\cite{kembhavi:cvpr2017} introduced TQA dataset, which aims to evaluate a system that contains multi-modal contexts and diverse topics in the scientific domain.
Similar datasets such as AI2D~\cite{kembhavi:eccv2016}, S{\tiny CIENCE}QA~\cite{lu:nips2022} are other large-scale TQA datasets with annotated lectures and explanations covering diverse science topics including natural science, social science, and language science.
Some other datasets such as GeoSQA~\cite{huang:geosqa2019}, MedQA~\cite{jin:arxiv2020}, MedMCQA~\cite{pal:pmlr2022} cover specific subjects such as geography and medicine.
A recently released TheoremQA dataset further covers textbook questions at the university level~\cite{chen:arxiv2023}.

\noindent \textbf{Methods.} \changed{From the technical perspective, TQA can be deemed as Visual Question Answering (VQA) in nature~\cite{dosovitskiy:iclr2020,gao:cvpr2019,gao:cvpr2022}.
Conventional VQA studies employ deep-learning models to encode the questions and images~\cite{antol:iccv2015,malinowski:iccv2015}.
Then the multi-modal information is fused together to comprehend the questions.
While there are significant similarities between VQA and TQA, we highlight some specific obstacles and limitations involved due to the distinct educational scenarios.
\begin{itemize}[leftmargin=*]
    \item 
\textbf{The subject-specific QA data is limited for training.}
Even though PLMs are frequently utilized as the backbone model, they are short of the subject knowledge.
To bridge the gap, some studies~\cite{ram:arxiv2021,xu:arxiv2022} fine-tune the contextual knowledge in the dataset as language modeling, which boosts the performance of TQA when only a small number of examples are available in the specific domains:
\begin{align*}
L_{\text{PreTrain}} = - \mathbb{E}_{P \in \mathcal{D}} \log P_{\theta} (P_{m}| Q, P_{\backslash m}),
\end{align*}
where $P_{m}$, $P_{\backslash m}$ represent the masked tokens and unmasked tokens respectively.
Some other studies combine diverse datasets in different subjects:
\begin{align*}
L_{\text{TQA}} = - \mathbb{E}_{(P, Q, A) \in \mathcal{D} \bigcup \Tilde{\mathcal{D}}} \log P_{\theta} (A| Q, P),
\end{align*}
where $\Tilde{\mathcal{D}}$ is augmented data set from different subjects.
$P_{\theta}$ is a VQA model trained with adversarial training to learn subject-invariant features~\cite{lee:mrqa2019}.
\item \textbf{The contexts may be presented in diverse formats.}
To understand the diagrams and tables, graph-based parsing methods are developed to extract the concepts from diagrams~\cite{kembhavi:eccv2016}.
Optical Character Recognition (OCR) is employed to identify the answers from the charts, which is further aligned with the questions~\cite{poco:cgf2017,kafle:cvpr2018}. 
These methods conduct pre-processing:
\begin{align*}
\tilde{P} = \mathcal{T}(P).
\end{align*}
Here, $\mathcal{T}$ is a set of off-the-shelf toolkits for calling. 
After that, $\tilde{P}$ will be deemed as the contexts of $Q$.
\item \textbf{Zero-shot issues frequently occur}.
TQA in education has a higher requirement for reasoning under zero-shot scenarios.
Recently, LLMs have shown impressive zero-shot capabilities in various NLP tasks, a number of studies~\cite{lu:nips2022,wang:aaai2024,wang_mtqa:arxiv2023} applied GPT-3 to solve the TQA.
To apply the full advantage of LLMs to image modality, they use caption models to translate visual information into language modality.
The paradigm of TQA systems becomes:
\begin{align*}
\tilde{P}= \text{Image2Text}(P); \quad A = \text{LLM}(Q, \tilde{P}).
\end{align*}
Liu~\etal~\cite{liu:arxiv2023} extended LLMs to multi-modalities, which results in a multi-modal model that connects a vision encoder and an LLM for general-purpose visual and language understanding:
\begin{align*}
A = \text{MLLM}(Q, P).
\end{align*}
More recent studies~\cite{lu:arxiv2023,chen:arxiv2023} develop compositional reasoning with LLMs as a planner that assembles a sequence of tools (e.g., \textit{Hugging Face} as image captioner, \textit{Github} as text detector, \textit{Web Search} as bing search, and \textit{Python} as program verifier.) to execute and answer the textbook questions.
We can formulate their methods as:
\begin{align*}
A = \text{LLM}(Q, P, \mathcal{T}).
\end{align*}
\end{itemize}
}

\ysadd{The above methods could effectively solve part of the challenges, but based on Table~\ref{tab:qg_datasets}, there is still room for improvement and we observe a severe performance gap of TQA tasks between the mainstream and the non-mainstream subjects such as geography and medicine, which is caused by the imbalanced collection of QA data.
This could be solved by constructing larger and wider scale of educational data.}

\subsubsection{Math Word Problem solving}

\changed{
Math Word Problem (MWP) solving is a typical category of QA specifically in mathematics subjects, aiming to convert a narrative description to an abstract expression:
\begin{align*}
    L_{\text{MWP}} = - \mathbb{E}_{(Q, E) \in \mathcal{D}} \log P_{\theta} (E| Q).
\end{align*}
Here, $Q$ and $E$ are a question and its corresponding math expression, which consists of a sequence of tokens and symbols, respectively.
This task is challenging because there remains a wide semantic gap in parsing human-readable words into machine-understandable logic.
}

\noindent \textbf{Datasets.} Dolphin-$18$K~\cite{huang:acl2016} is an early large-scale MWP solving dataset, which contains over $18,000$ annotated math word problems of elementary mathematics.
DRAW-$1$K~\cite{upadhyay:arxiv2016} is introduced as a testbed for algebra word problem solvers.
Math$23$K~\cite{wang:emnlp2017} is a Chinese MWP solving dataset.
Subsequently, MathQA~\cite{amini:naacl2019} and ASDiv~\cite{miao:acl2020} further extend the scale and diversity of the existing datasets, which makes it particularly suited for usage in training deep learning models to solve MWP problems.
Cobbe~\etal~\cite{cobbe:arxiv2021} introduced a more challenging dataset GSM$8$K, which is complicated in high linguistic diversity.
IconQA~\cite{lu:nips2021} extends MWP solving to abstract diagram understanding and comprehensive cognitive reasoning by involving diagrams in problems.
Moreover, T{\small AB}MWP~\cite{lu2022dynamic} includes tables to enhance the difficulty of the MWPs, which requires the table parsing and understanding capabilities~\cite{lu2022dynamic}.
As we can see, there is a set of MWP datasets that examine various reasoning capabilities.
With the development of this task, the evaluated datasets become increasingly complicated in linguistic diversity and multi-modalities. 

\noindent \textbf{Methods.} 
MWP solving is a popular NLP application in the education domain for granted since it can be utilized as a teaching assistant in math.
The investigation of MWP solving can be traced back to the $1960$s~\cite{zhang:tpami2020}.
Early work leverages manually crafted rules and schemas for pattern matching, where human intervention is heavily relied on.
Subsequently, various strategies of feature engineering and statistical learning are proposed to boost the performance.
With the development of techniques and the emergence of large-scale datasets, deep learning-based methods have been widely applied to MWP solving.
The majority of the methods follow the paradigm of machine translation~\cite{wang:emnlp2017}, which first encodes the narratives, then decodes the mathematical expression token as the output sequence.
Compared with machine translation, there are some distinct solutions that have been proposed due to the structured outputs of MWPs. 
\changed{
\begin{itemize}[leftmargin=*]
\item \textbf{Some MWPs are complex in logic.} To improve the generation of complex solution expressions, some methods are proposed to formulate a mathematical expression as an abstract tree~\cite{wang:emnlp2018,chiang:naacl2019,liu:emnlp2019,xie:ijcai2019}.
To improve the understanding of the narratives, some studies focus on improving the encoding of the narratives by re-organizing the textual contexts with structure~\cite{zhang:acl2020,lin:aaai2021}, which may increase computation complexity as a drawback. 
In this case, graph networks are frequently involved in:
\begin{align*}
    \mathcal{E} = P_{\theta}(E |Q) & = \text{GNN}(\mathcal{Q}).
\end{align*}
Here, equations and questions can be represented as dependency trees $\mathcal{Q}$ and abstract syntax trees $\mathcal{E}$, respectively.
\item \textbf{There is a lack of annotated expressions in the MWPs.} 
Due to the weak supervision caused by the lack of annotated expressions, multiple studies~\cite{wang:emnlp2018,hong:aaai2021} model the MWP solving task as follows:
\begin{align*}
    L_{\text{MWP}} = - \mathbb{E}_{(Q, A) \in \mathcal{D}} \log \sum_{E} P(A| E) P_{\theta} (E| Q).
\end{align*}  
Also, some researchers~\cite{zhang:ijcai2020} apply knowledge distillation to MWP solving tasks, which learns a smaller model from a large pre-trained generic model but may also include some unexpected noise.
\item \textbf{Zero/few-shot complex MWPs in real-world scenarios.}
MWP solving usually encounters problems with complex reasoning steps under zero-shot.
Chain-of-Thought (CoT) is a typical prompting strategy that is meant to decompose a complicated MWP into sub-questions such that LLMs could conduct reasoning step by step~\cite{Wei:arxiv2023}.
Follow-up methods improve it by designing more interactive prompting or constructing more efficient demonstrations~\cite{Wang:arxiv2023, Zhang:arxiv2022, Zhou:arxiv2022, Zheng:arxiv2023,lu:arxiv2023}.
Similar to the TQA tasks, researchers~\cite{chen:tmlr2023} leverage LLMs as an agent~\cite{guo:arxiv2023} and delegate the computation to a program interpreter, which can be conducted by the LLMs and executed by a Python interpreter.
\end{itemize}
}
\ysadd{We observe the performance of MWP solving has a large variance across different datasets in Table~\ref{tab:qg_datasets}.
With the increasing complexity of the modalities and questions, the tasks become more challenging. Hence, more advanced and powerful methods should be explored to improve the performance.}







\subsection{Question Construction}

\begin{table*}[t!]
\centering
\resizebox{\textwidth}{!}{
\begin{tabular}{l c c c c | c c c}
        \toprule  
        Dataset & Science subject & Source & Q\&A type & \#Q & Top 1 & Top 2 & Top 3 \\
        \midrule
        \multirow{2}{*}{SciQ$\vartriangle$} & \multirow{2}{*}{science} & online science & normal &   \multirow{2}{*}{$13.7$K}&$37.2$/$-$/$-$~\cite{bulathwela2023scalable}&$36.7$/$-$/$-$~\cite{bulathwela2023scalable}&$29.4$/$17.2$/$29.5$~\cite{zhu2022unsupervised} \\
        & & textbooks& MC\&extractive &&EduQG~\cite{bulathwela2023scalable}&Leaf~\cite{lopez2021simplifying}&DDS+ST~\cite{zhu2022unsupervised} \\
        \hdashline
        \multirow{2}{*}{RACE$\vartriangle$} & \multirow{2}{*}{linguistic} & secondary school & normal\&cloze &  \multirow{2}{*}{$100$K}&$39.8$/$-$/$34.0$~\cite{chung2020bert}&$36.5$/$-$/$31.6$~\cite{chung2020bert}&$24.7$/$15.1$/$23.8$~\cite{zhu2022unsupervised} \\
        & & English exams & MC\&extractive & &BDG~\cite{chung2020bert}&GPT&DDS+ST~\cite{zhu2022unsupervised}\\
        \hdashline
        \multirow{3}{*}{LearningQ$\vartriangle$} &  computing/& \multirow{2}{*}{ online education}& \multirow{2}{*}{normal}  & \multirow{3}{*}{$231$K}&\multirow{2}{*}{$24.7$/$8.7$/$27.4$~\cite{chen2018learningq}}&\multirow{2}{*}{$19.8$/$6.4$/$23.1$~\cite{chen2018learningq}}&\multirow{2}{*}{$0.4$/$3.0$/$6.5$~\cite{chen2018learningq}} \\
        &  science/business/ & \multirow{2}{*}{platform}&\multirow{2}{*}{-} & &\multirow{2}{*}{Attention~\cite{du2017learning}}&\multirow{2}{*}{Seq2Seq~\cite{luong2014addressing}}&\multirow{2}{*}{H\&S~\cite{heilman2010good}} \\
        & humanities/math&  & &  \\
        \hdashline
        \multirow{2}{*}{KHANQ$\vartriangle$} & \multirow{2}{*}{science} & online education & normal  & \multirow{2}{*}{$1$K}& $25.6$/$12.7$/$27.6$~\cite{gong2022khanq}& $25.1$/$12.7$/$26.1$~\cite{gong2022khanq}& $20.2$/$10.8$/$22.0$~\cite{gong2022khanq} \\ 
        & & platform &extractive & &T5~\cite{raffel2023exploring}& BART~\cite{lewis2019bart}& UniLM~\cite{dong2019unified}\\
        \hdashline
        \multirow{3}{*}{EduProbe$\vartriangle$} &history/geography/& \multirow{2}{*}{grades 6-12}&\multirow{2}{*}{normal}&\multirow{3}{*}{$3.5$K}& \multirow{2}{*}{$43.0$/$50.3$/$66.8$~\cite{maity2023harnessing}}& \multirow{2}{*}{$36.5$/$44.3$/$66.6$~\cite{maity2023harnessing}}& \multirow{2}{*}{$33.6$/$41.7$/$64.0$~\cite{maity2023harnessing}}\\
        &economics/science/&\multirow{2}{*}{textbooks}&\multirow{2}{*}{extractive}&&\multirow{2}{*}{T5~\cite{raffel2023exploring}}& \multirow{2}{*}{BART~\cite{lewis2019bart}}& \multirow{2}{*}{MBART~\cite{liu2020multilingual}}\\
        &environmental studies&&&&&&\\
        \hdashline
        \multirow{4}{*}{EduQG$\vartriangle$} & politics/physiology/ & \multirow{4}{*}{online textbooks} & \multirow{3}{*}{normal\&cloze}  & \multirow{4}{*}{$3$K}& \multirow{3}{*}{$33.7$/$45.8$/$-$~\cite{hadifar2023diverse}}& \multirow{3}{*}{$15.4$/$29.7$/$34.3$~\cite{hadifar2023eduqg}}& \multirow{4}{*}{-} \\ 
        & biology/business/&  &\multirow{3}{*}{MC\&extractive} & &\multirow{3}{*}{DCS~\cite{hadifar2023diverse}} &\multirow{3}{*}{T5~\cite{raffel2023exploring}}\\
        & law/sociology/&  & & & \\
        & psychology/history&  & & & \\
        \hdashline
        \multirow{2}{*}{MCQL$\dagger$} & biology/physics/ & \multirow{2}{*}{website} & normal  & \multirow{2}{*}{$7.1$K}&$35.4$/$-$/$-$~\cite{liang2018distractor}&$34.5$/$-$/$-$~\cite{liang2018distractor}&$33.6$/$-$/$-$~\cite{liang2018distractor} \\
        & chemistry&  &MC &&LR+RF~\cite{liang2018distractor}&LR+LM~\cite{liang2018distractor}&RF~\cite{liang2018distractor}\\
        \hdashline
        \multirow{3}{*}{Televic$\dagger$} & math/health/ & \multirow{2}{*}{ online education} & \multirow{2}{*}{normal} & \multirow{3}{*}{$62$K}& \multirow{2}{*}{$-$/$58.8$/$16.4$~\cite{bitew2023distractor}}& \multirow{2}{*}{$85.8$/$28.9$/$50.1$~\cite{bitew2022learning}}& \multirow{2}{*}{$-$/$27.8$/$36.6$~\cite{bitew2023distractor}} \\
        & history/geography/& \multirow{2}{*}{platform} &\multirow{2}{*}{MC} &  & \multirow{2}{*}{ChatGPT}& \multirow{2}{*}{DQ-SIM~\cite{bitew2022learning}}& \multirow{2}{*}{mT5~\cite{xue2020mt5}}\\
        & science&  & & & \\
        \bottomrule
    \end{tabular}
}
\caption{Comparison of QG and DG datasets. The format of the question is categorized as either normal (e.g., interrogative sentence.) or cloze (e.g., fill-in-the-blank.), and the answer is categorized as either multiple-choice (MC) or extractive (parts of the text are considered as the answer)~\cite{hadifar2023eduqg}. 
\changed{The evaluation metrics are BLEU-4/METEOR/ROUGE-L$\vartriangle$ for QG and MAP (Mean Average Precision)/GDR (Good Distractor Rate)/NDR (Non-sense Distractor Rate, lower is better)$\dagger$ for DG}
}
\label{tab:qg_datasets}
\vspace{-0.5cm}
\end{table*}



\begin{figure}[th!] 
	\centering
	\includegraphics[width=0.5\textwidth]{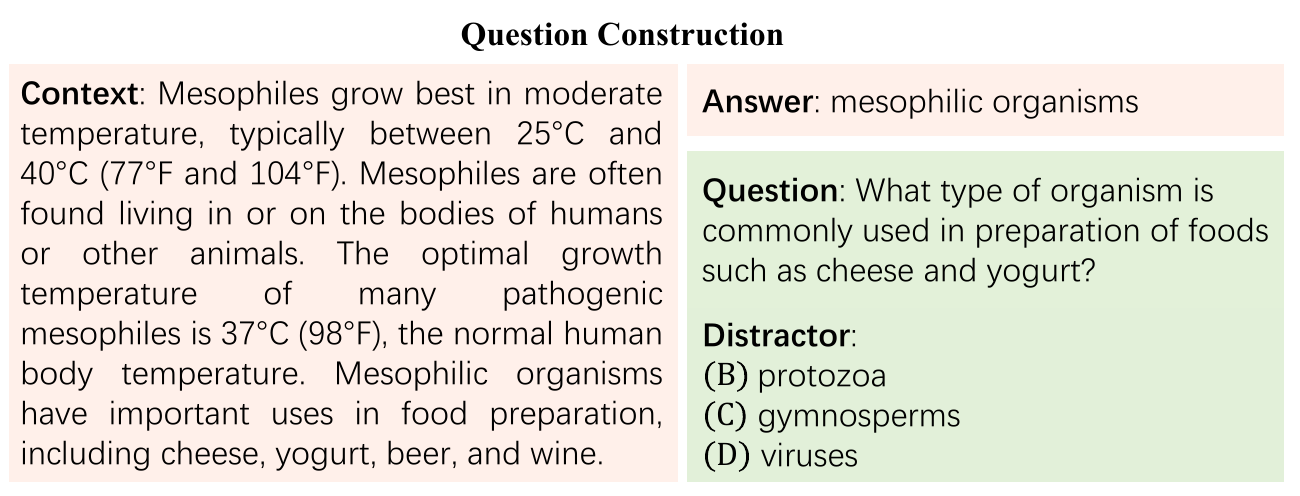}
	\caption{The examples of Question Construction. 
The displayed examples are extracted from SciQ.
    }
	\label{fig:eduqc}
\end{figure}

Question Construction (QC) aims to automatically construct questions from a given context, which performs a significant role in education~\cite{das2021automatic}. 
Usually, Multiple Choice Questions (MCQ) are common question types for a test quiz.
The construction of MCQ consists of two parts: Question Generation (QG) and Distractor Generation (DG). 
The former generates questions based on a given context, while the latter generates distractors to complement the correct ones. 
In Figure~\ref{fig:eduqc}, we display a typical example of QG and DG in the education domain.
In this section, we introduce the relevant datasets and methods for QG and DG in detail.

\subsubsection{Question Generation} 
\changed{
The goal of QG is to automatically generate a question from a given sentence or paragraph.
The challenge lies in identifying the key statement based on the context with or without the utilization of the answer (namely answer-aware QG or answer-agnostic QG~\cite{dugan2022feasibility}) and generating a question based on the statement, which can be described as follows:
\begin{align*}
L_{\text{QG}} = -\mathbb{E}_{(C, [A], Q) \in \mathcal{D}} \log P_{\theta}(Q \mid C, [A])
\end{align*}
Here $C$ and $Q$ are context and the generated question, respectively.
$[A]$ is the answer which is optional to the input.
}


\noindent \textbf{Datasets.} 
QG and QA are dual tasks~\cite{tang2017question}. 
They both require reasoning between questions and answers. 
As a result, some MCQ datasets initially designed for QA tasks, including SciQ~\cite{welbl2017crowdsourcing} and RACE~\cite{lai2017race}, are also leveraged for research on QG~\cite{wu2023towards,jia2021eqg,steuer2022investigating,zhao2022educational}.
Besides, there are some datasets particularly constructed for QG, such as LearningQ~\cite{chen2018learningq}, KHANQ~\cite{gong2022khanq}, EduQG~\cite{hadifar2023eduqg} and EduProbe~\cite{maity2023harnessing}.
These datasets extend across various subjects (e.g., science, medicine) and various education levels (e.g., preliminary school, middle school, and university.), the details of which are displayed in Table~\ref{tab:qg_datasets}.
Regarding these datasets, LearningQ is collected from instructor-crafted learning questions, which contain cognitively demanding documents for QG that require reasoning, but the absence of answers limits its application to answer-aware QG. 
KHANQ and EduProbe provide (context, textual cue, question) triples.
EduQG is a high-quality MCQ dataset generated by educational experts and questions are linked to their cognitive complexity, which makes this dataset move closer to reality.


\noindent \textbf{Methods.}
Early methods for general QG primarily relied on rule matching~\cite{kunichika2004automated}. 
Du~\etal~\cite{du2017identifying} first applied the Seq2Seq model with attention to automatic QG. 
Built upon this framework, methods such as integrating linguistic features~\cite{naeiji2022question}, training a multi-modal QG model~\cite{wang2023multiqg}, leveraging multi-task learning~\cite{zhou2019multi} and reinforcement learning~\cite{chen2019natural} have also been used to optimize the QG models. 
With the advent of language models, the BERT model and GPT-3 have been applied to QG~\cite{chan2019recurrent,wang2022towards}.
Besides, regarding answer information in the QG system, researchers proposed to encode the answer together with the context~\cite{yuan2017machine}, utilize the position information of the answer in the context~\cite{ma2020improving}, and achieve answer-aware QG, while using summaries of the texts as input can serve as a strategy for answer-agnostic QG as it can help the modeling of topic salience, which denotes the importance of specific words or phrases in the context~\cite{dugan2022feasibility}.
Importantly, there are some unique challenges when we adopt QG systems to the education domain. 
\changed{
\begin{itemize}[leftmargin=*]
\item \textbf{The generated questions ignore the difficulty stratification.} 
Some general methods lack an effective way to control the difficulty of the question, which is vital for improving the efficiency of education. The difficulty-controllable QG task can be described as:
\begin{align*} 
L_{\text{QG}} = -\mathbb{E}_{(C, A, Q, D) \in \mathcal{D}} \log P_{\theta}(Q \mid C, A, D)
\end{align*}
Here $D$ represents the difficulty levels, which can be formulated in several ways: whether a well-trained QA model can answer it~\cite{gao2018difficulty}, the number of inference steps~\cite{cheng2021guiding}, and the educational paradigms like Item Response Theory~\cite{uto2023difficulty} or Bloom's Taxonomy~\cite{forehand2005bloom} that categorize learning objectives into cognitive, affective, and psychomotor domains.
Recently, Hwang~\etal~\cite{hwang2024towards} applied few-shot prompting of GPT-4 to generate questions aligned with Bloom's Taxonomy~\cite{forehand2005bloom}. Lee~\etal~\cite{lee2024few} and Maity~\etal~\cite{maity:fire2024} also designed an effect prompt for the QG system to cover a wide range of difficulty levels.
\item \textbf{The generated questions show weak relevance to the syllabus.} 
Aligning generated questions to the syllabus can be beneficial for identifying the focus of the test quiz, which can be achieved by including a selection step for content in general QG methods. Multiple studies~\cite{steuer2021not,hadifar2023diverse} train binary classifiers or ranking models to judge whether the generated question is related to concepts or pedagogical contents with the objective of maximizing both relevance and diversity. 
\begin{align*} 
P_{\phi}(K\mid Q, C) = \text{Classifier}(Q, C, K),
\end{align*}
where $K$ represents certain keywords or topics of the syllabus. 
Besides, Xiao~\etal~\cite{xiao2023evaluating} generated questions that align with given topic keywords using the plug-and-play language models~\cite{dathathri2019plug}.
\item \textbf{The generated questions should be customized to students.} 
QG systems for customized education focus on generating personalized questions for students. 
Srivastava~\etal~\cite{srivastava2021question} developed a knowledge-tracking model that uses students' answer histories to predict their performance on new questions, fine-tuning it with an autoregressive language model.
Subsequently, Wang~\etal~\cite{wang2023difficulty} introduced the few-shot knowledge tracking model to predict question difficulty by incorporating sequences of student states and questions.
\begin{align*} 
L_{\text{QG}} = -\mathbb{E}_{(C, S, Q, D) \in \mathcal{D}} \log P_{\theta}(Q \mid C, S, D)
\end{align*}
Here $D$ is the difficulty level of the question. $S$ is the state of the students, which is represented as a temporally-evolving sequence of questions and
their responses~\cite{srivastava2021question}. 
Particularly, given a student's responses as $a$ (correct$\langle Y \rangle $/incorrect$\langle N \rangle$) to $m$ questions as $q$, the state $s$ can be formulated as $s = \{ q_1, a_1, \ldots q_m, a_m \}$.
\end{itemize}
}
\ysadd{Even though various methods have been proposed, question generation is still a difficult task to ensure its alignment to the real world.
More studies focusing on developing syllabus-guided and customized questions are highly-anticipated.
}

\subsubsection{Distractor Generation} 

\changed{As we mentioned above, DG is an associated component for question construction, which aims to generate distractors with the given context, question, and correct answer in a MCQ. 
A good distractor should be relevant to the context, grammatically coherent to the interrogative of the question, and deceptive as a false answer~\cite{welbl2017crowdsourcing}.
The task is generally formulated as follows:
\begin{align*}
L_{\text{DG}} & = -\mathbb{E}_{(C, Q, A, O) \in \mathcal{D}} \log P_{\theta}(O  \mid C, Q, A) 
\end{align*}
Here, $C$, $Q$, $A$, and $O$ are context, question, answer, and distractors, respectively. 
}

\noindent \textbf{Datasets.} 
The aforementioned MCQs for QG tasks such as SciQ and RACE can also be leveraged for DG tasks. 
Specifically, the distractors in SciQ are designed by crowd workers with reference~\cite{welbl2017crowdsourcing}. 
Instead of providing entities and concepts as distractors, the distractors of RACE are at the sentence level~\cite{qiu2020automatic}, which are collected from the secondary school English exams. 
Alongside the datasets mentioned before, there are other MCQ datasets especially designed for DG. 
The data of MCQL~\cite{liang2018distractor} is crawled from the web and provides distractors that are either phrases or sentences, covering multi-subjects. 
Besides, Televic~\cite{bitew2022learning} is collected through Televic Education's platform with distractors manually crafted by experts across multiple languages and domains.

\noindent \textbf{Methods.} Early studies for DG are rule-based and feature-based, which rely on linguistic rules~\cite{mitkov2003computer} and specific features like word frequency~\cite{brown2005automatic} or diverse similarity measurement between distractors and correct answers~\cite{liang2018distractor}. With the rapid development of deep learning, the construction of distractors can be divided into two lines of mainstream: generation-based and ranking-based categories~\cite{wang2023multi}. Generation-based methods automatically generate distractors token by token and ranking-based methods regard DG as a ranking task for a pre-defined distractor candidate set.
Hence, $O$ could be either generated or selected via generation-based or ranking-based models.
Noting that there are some key challenges of DG to make a good distractor.
\begin{itemize}[leftmargin=*]
\item \textbf{The produced distractors are irrelevant to the questions.} 
Some studies~\cite{bitew2023distractor, wang2023multi, yu2024enhancingdistractorgenerationmultiplechoice} focus on generating distractors that are consistent with the context or the answer.
Gao~\etal~\cite{gao2019generating} proposed a hierarchical model with dual attention for sentence and word relevance and irrelevant information filtering to generate distractors that are semantically consistent with and traceable in the context. Ding~\etal~\cite{ding2024can} propose a multi-modal DG model integrating contrastive learning to ensure consistency and have high relevance to the contexts. 
\item \textbf{The distractors lack strong distractive effectiveness.} 
A good distractor should be misleading~\cite{alhazmi2024distractor}, with high similarity to the correct answer and distinguishability from other distractors. Taslimipoor~\etal~\cite{taslimipoor2024distractor} proposed a two-step approach using transfer learning to generate correct answers and distractors simultaneously, followed by clustering to classify and remove duplicates to avoid excessive similarity among distractors. Qu~\etal~\cite{qu2024unsupervised} proposed an unsupervised DG framework using distillation from LLMs and contrastive decoding to avoid excessive similarity between generated distractors.
\end{itemize} 

\ysadd{Our investigation reveals that DG, as a significant component of QG, has not been fully explored. More discussion on constructing DG datasets, improving evaluation strategies, and generating ``good'' distractors should be learned.}


\subsection{Automated Assessment}

\begin{table*}[t!]
\centering
\resizebox{\textwidth}{!}{
\begin{tabular}{l l c c c c c c c}
        \toprule  
        Dataset & Essay Types & Source & Language & Exp & \#Q & Top 1 & Top 2 & Top 3 \\ 
        \midrule
        \multirow{2}{*}{CLC-FCE$\vartriangle$}   & \multirow{2}{*}{A/N/C/S/L}  & non-native:  & \multirow{2}{*}{English} & \multirow{2}{*}{holistic} & $1$K & $61.7$~\cite{schmalz:ccl2021} & $61.5$~\cite{schmalz:ccl2021} & $61.5$~\cite{schmalz:ccl2021} \\ 
        & & ESOL test & & & & Two-stream~\cite{schmalz:ccl2021} & Autom~\cite{schmalz:ccl2021} & Text-only~\cite{schmalz:ccl2021} \\
        \hdashline
        \multirow{2}{*}{ASAP++$\vartriangle$}  & \multirow{2}{*}{A/R/N} & native:  & \multirow{2}{*}{English} & \multirow{2}{*}{holistic} & $17$K & $71.0$~\cite{chen:aied2024} & $69.8$\cite{li:acl2024} & $53.0$~\cite{chen:acl2023} \\ 
        & & grades 7-10 & & & & MTAA~\cite{chen:aied2024} & Li\&Ng~\cite{li:acl2024} & Chen\&Li~\cite{chen:acl2023} \\
        \hdashline
        \multirow{2}{*}{ICLE++$\dagger$} & \multirow{2}{*}{A} & non-native: & \multirow{2}{*}{English} & organization/... & $4$K & $13.4/12.2/27.9$\cite{li:naacl2024} & $18.9/14.4/23.4$\cite{li:naacl2024} & $8.2/6.0/14.1$\cite{} \\ 
        & & undergraduate exam & & /persuasiveness & & PMAES\cite{li:naacl2024} & Kumar\cite{kumar:naacl2022} & Uto\cite{uto:coling2020} \\
        \hdashline
        \multirow{2}{*}{ELLIPSE$\vartriangle$} & \multirow{2}{*}{A} & native: & \multirow{2}{*}{English} & \multirow{2}{*}{holistic} & $4$K & $68.0$~\cite{chen:aied2024} & $67.0$~\cite{chen:aied2024} & $66.0$~\cite{chen:aied2024} \\
        & & grades $8-12$ & & & & MTAA~\cite{chen:aied2024} & MTL-Vanilla~\cite{chen:aied2024} & RoBERTa \\
        \hdashline
        \multirow{2}{*}{HSK$\vartriangle$}  & \multirow{2}{*}{A/N} & non-native:  & \multirow{2}{*}{Chinese}  & \multirow{2}{*}{holistic} & $10$K & $71.4$\cite{yang:as2023} & $70.4$\cite{yang:as2023} & - \\ 
        & & HSK test & & & & DNN\cite{sun:entropy2022} & LogReg\cite{wang:ccl2021} & - \\
        \bottomrule
    \end{tabular}
}
\caption{Comparison of AES datasets.
The essay types contain: argumentative (A), response (R), narrative (N), comment (C), suggestion (S) and letter (L).
``Exp'' denotes the type of scores annotated.
``\#Q'' denotes the number of questions.
\changed{We display most datasets with holistic Score Accuracy$\vartriangle$ except that ICLE++ is displayed with prompt adhere/organization/cohesion Score Accuracy$\dagger$.
}
}
\label{tab:aes_dataset}
\vspace{-0.5cm}
\end{table*}

\begin{figure}[th!] 
	\centering
	\includegraphics[width=.5\textwidth]{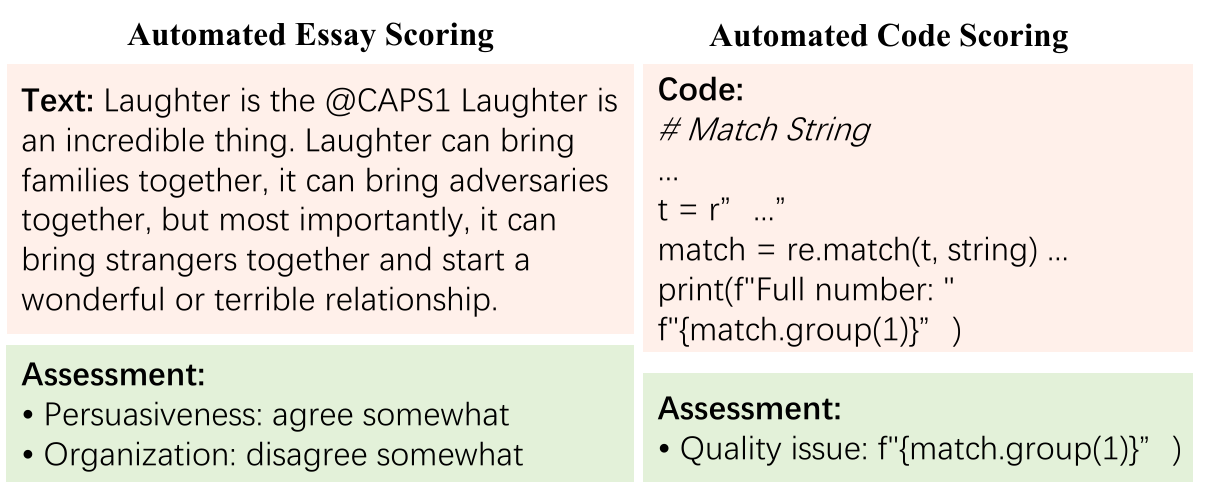} 
	\caption{The examples of AES and ACS tasks.
 The displayed examples are extracted from ASAP dataset and paper~\cite{tigina:arxiv2023}, respectively.
 }
	\label{fig:eduas}
\end{figure}

Automated assessment is a widely-investigated task in the education domain, which is helpful in reducing the burden of teachers in various writing tasks.
Regarding different objects to be scored, different criterion is considered.
For instance, in Figure~\ref{fig:eduas}, scoring an essay in a language curriculum requires the measurement of particular essay quality such as persuasiveness and organization.
For a piece of code in the program curriculum, it is more important to measure the correctness of the code, the completeness of the comments, and quality issues.
Therefore, we discuss the techniques for Automated Essay Scoring (AES) and Automated Code Scoring (ACS) in the following subsections.

\subsubsection{Automated Essay Scoring}

\changed{
AES aims to develop a system to score an essay automatically, where the written text is given as the input and the system summarizes its quality as a score:
\begin{align}
    L_{\text{AES}} = -\mathbb{E}_{(E, s) \in \mathcal{D}}P_{\theta}(s | \mathbf{f}(E)),
\label{eq:l_aes}
\end{align}
where $s$ is a holistic score and $\mathbf{f}(E)$ is a set of features derived from an essay $E$. 
The challenge of this task lies in understanding the content of the essay by modeling the different facets of the text.
}

\noindent \textbf{Datasets.} The Cambridge Learner Corpus-First Certificate in English exam (CLC-FCE)~\cite{yannakoudakis:acl2011} is a widely used dataset for AES tasks.
Usually, a dataset is annotated with the prompt of the essay, which indicates the topic of the essay.
Moreover, it provides both the holistic score and the manually tagged linguistic error types such that it makes possible to evaluate the instructional feedback of the developed AES systems.
The Automated Student Assessment Prize (ASAP) is a Kaggle competition released for AES.
A similar dataset ELLIPSE~\cite{alex:kaggle2022} contains essays from English language learners of the diverse difficulty levels.
Instead of holistic scores, scores along different dimensions of essay quality are needed.
International Corpus of Learner English (ICLE++)~\cite{li:naacl2024} is a dataset containing essays annotated with the perspectives of structure organization, thesis clarity, prompt adherence, and argument persuasiveness.
There are a few AES datasets, which are designed for other languages such as Chinese~\cite{wang:ccl2021}.
We display the details of the above datasets in Table~\ref{tab:aes_dataset}.

\noindent \textbf{Methods.} 
The majority of the AES studies usually predict a single score for an essay.
In Equation~(\ref{eq:l_aes}), $P_{\theta}(\cdot)$ can be formulated as a regression task, a classification task, a ranking task or their hybrid~\cite{yang:emnlp2020,wang:acl2023}. 
A series of features such as length-based features~\cite{zesch:bea2015}, lexical features~\cite{phandi:emnlp2015,zesch:bea2015}, prompt-relevant features~\cite{louis:bea2010}, readability features~\cite{zesch:bea2015}, syntactic features~\cite{zesch:bea2015,chen:emnlp2013} and so on, are first extracted from the essay, then off-the-shelf learning algorithms are leveraged to fit the annotated scores~\cite{yamaura:aied2023}. 
Besides the hand-crafted features, $\mathbf{f}(E)$ can be modules automatically capturing the document structure~\cite{uto:coling2020,sharma:ecml_pkdd2021}.
In addition, there are some focuses existing in the AES tasks for education. 

\changed{
\begin{itemize}[leftmargin=*]
\item \textbf{The singular evaluation of the essays lacks comprehensiveness}.
Multiple dimensions are considered in a few studies such as cohesion, syntax, vocabulary, phraseology, grammar, and conventions\cite{persing:acl2013,persing:acl2014}, which are able to construct more diverse evaluation metrics.
To enrich the evaluation of the AES task, researchers utilize LLMs to provide a more comprehensive and explainable evaluation of an essay~\cite{chen:aied2024}.
With the trait-specific score, the AES system becomes multi-tasking learning:
\begin{align*}
    L_{\text{AES}} = -\mathbb{E}_{(E, s) \in \mathcal{D}} \sum_{s^k \in s}P_{\theta}(s^k | \mathbf{f}(E)),
\end{align*}
where $s^k$ measures different aspects of the essays.
\item \textbf{The sparse annotation of the essays in domain-specific prompts}. 
Cross-prompt AES aims to rate unseen target essays with annotated source prompts.
Multiple studies~\cite{jiang:acl2023,jiang:www2021,chen:acl2023} design neural networks decoupling the domain-dependent and domain-independent features, to facilitate the models to learn transitions ($s \rightarrow t$) from different domains: $t = s \times T^\intercal$,
where $T$ is a matrix modeling the transition from source prompts to target prompts.
\end{itemize}
}
\ysadd{AES also faces the unsolved issues of object-oriented and scenarios-oriented scoring. Specifically, the essay scores of K12 learners, second-language learners and businessmen are different.
Maybe a more fine-grained scoring paradigm or framework should be developed.}

\subsubsection{Automated Code Scoring}

Similar to AES, ACS grades the score of a code snippet on various dimensions.
Compared with essay grading, code grading is more complicated in understanding the long-dependency inside a code snippet and its execution effect could be a vital factor to judge.



\noindent \textbf{Datasets.}
We notice that there are few publicly-available ACS datasets due to the privacy issue.
The source data contain submissions in programming courses that can be collected on the academy platforms such as JetBrains Academy, MOOC, tutoring sessions and so on~\cite{birillo:cse2022,birillo:arxiv2023,tigina:arxiv2023}.

\noindent \textbf{Methods.} 
Code can be scored from different objectives, which is vital in judging the quality of a code snippet.
Intuitively, by executing on a set of test cases, the execution results are supposed to be the same as the excepted results.
\begin{align*}
    \{\text{pass}, \text{fail}\} =\text{Compiler}(C, \text{case}),
\end{align*}
where $C$ is a piece of code snippet.
In this way, many dynamic analysis tools for detecting code quality issues are developed, such as \textit{AutoStyle}~\cite{roy:its2016}, \textit{WebTA}~\cite{ureel:tscse2019}, \textit{Refactor Tutor}~\cite{keuning:tscse2021} and \textit{Hyperstyle}~\cite{birillo:cse2022}. 
However, it is challenging to score a code snippet without executing it.

\begin{itemize}[leftmargin=*]
\item \textbf{It is hard to identify good features and scoring algorithms.}
The static features include complexity, comments, number of lines, number of operators, number of operands, and so on. 
The static analysis approaches leverage these features to identify the quality of code~\cite{lajis:icsca2018}. 
Machine learning algorithms also have been applied to ACS.
Clustering, breadth-first search, and neural network approaches are utilized to track the logic errors in code~\cite{glassman:tochi2015,talbot:sce2020,von:ie2018}.
Besides, Queiros~\etal~\cite{queiros:aicse2012} proposed hybrid approaches combining static and dynamic approaches, which can be formulated as follows:
\begin{align*}
    L_{\text{ACS}} = -\mathbb{E}_{(C, s) \in \mathcal{D}} P_{\theta}(s | \mathbf{f}(C)),
\end{align*} 
where $C$ is the code snippet and $s$ is the annotated score for supervised training.
\item \textbf{Implicitly encode the code snippet via distributed representation.}
Instead of directly scoring a code snippet based on the extracted features, there are a series of studies focusing on evaluating the contextual embedding of a code snippet, which benefits measuring the distance between the evaluated snippet and the correct snippet for ACS~\cite{wang:mlkdd2023}.
The embedding of variable and method identifiers utilizes local and global context\cite{allamanis:fse2015}, Abstract Syntax Trees (AST)~\cite{zhang:icse2019}, and AST paths~\cite{alon:popl2019,hellendoorn:iclr2019}.
Later, Feng~\etal~\cite{feng:arxiv2020} and Kanade~\etal~\cite{kanade:pmlr2020} proposed CodeBERT and CuBERT, respectively. 
They model the code beyond simple tokenization: 
\begin{align*}
    L_{\text{PreTrain}} = -\mathbb{E}_{C \in \mathcal{D}} P_{\theta} (C_{m}| C_{\backslash m}).
\end{align*}
Similar to language models, the pre-training aims to learn the pattern of code language.
\end{itemize} 

\ysadd{Even though a great number of systems have been developed for ACS, there still remains an unsolved question to develop adaptive and fine-grained scoring paradigms for ACS in various scenarios.}


\subsection{Error Correction}

\begin{table*}[t!]
\centering
\small
\resizebox{\textwidth}{!}{
\begin{tabular}{l c c c c | c c c}
        \toprule  
        Dataset & Language & Source & Exp & \#Q & Top 1 & Top 2 & Top 3 \\
        \midrule
        \multirow{2}{*}{LANG-8} & Multi & \multirow{2}{*}{LANG-8 Website} & \multirow{2}{*}{\xmark} & \multirow{2}{*}{$1$M} &\multirow{2}{*}{-} &\multirow{2}{*}{-} &\multirow{2}{*}{-} \\ 
         & (80 languages) & & & & & & \\
        \hdashline
        \multirow{2}{*}{CoNLL-2014$\ast$} & \multirow{2}{*}{En} & \multirow{2}{*}{NUCLE~\cite{Dahlmeier_Ng_Wu_2013}} & \multirow{2}{*}{\cmark} & \multirow{2}{*}{$58$K} &$72.8$~\cite{omelianchuk2024pillars}&$72.1$~\cite{omelianchuk2024pillars}&$69.5$~\cite{omelianchuk2024pillars} \\
        & & & & &M-VOTING~\cite{omelianchuk2024pillars} &GRECO~\cite{qorib2023system} &ESC~\cite{qorib2022frustratingly} \\
        \hdashline
        \multirow{2}{*}{BEA-2019$\ast$} & \multirow{2}{*}{En} & NUCLE, FCE, W\&I,& \multirow{2}{*}{\cmark} & \multirow{2}{*}{$686$K} &$81.4$~\cite{omelianchuk2024pillars}&$80.8$~\cite{omelianchuk2024pillars}&$79.9$~\cite{omelianchuk2024pillars} \\ 
        &  & LOCNESS, LANG-8 & & &M-VOTING~\cite{omelianchuk2024pillars} &GRECO~\cite{qorib2023system} &ESC~\cite{qorib2022frustratingly} \\
        \hdashline
        \multirow{2}{*}{CCTC$\ast$} & \multirow{2}{*}{Zh} & Texts written by & \multirow{2}{*}{\cmark} & \multirow{2}{*}{$31$K} &$29.5/23.0$~\cite{wang2022cctc} &$22.3/16.9$~\cite{wang2022cctc} &$15.77/15.26$~\cite{wang2022cctc} \\
        & & native speakers& & &SpellGCN~\cite{Cheng_Xu_Chen_2020} &GECToR~\cite{Omelianchuk_2020} &SpellGCN~\cite{Cheng_Xu_Chen_2020}  \\
        \hdashline
        \multirow{2}{*}{FCGEC$\ast$} & \multirow{2}{*}{Zh} & \multirow{2}{*}{Chinese examinations} & \multirow{2}{*}{\cmark} & \multirow{2}{*}{$41$K} & $51.3$~\cite{wang2024lm} & $45.5$~\cite{wang2024lm} & $45.1$~\cite{wang2024lm} \\
        & & & & &LM-Combiner~\cite{wang2024lm} &STG~\cite{Xu_Wu_Peng_2022} &GPT2~\cite{Radford_Wu_Child_Luan_2018} \\
        \hdashline
        \multirow{2}{*}{FlaCGEC$\ast$} & \multirow{2}{*}{Zh} & \multirow{2}{*}{HSK~\cite{Baoli_2011}} & \multirow{2}{*}{\cmark} & \multirow{2}{*}{$13$K} &$64.9$~\cite{FlaCGEC} &$48.2$~\cite{FlaCGEC} &$37.3$~\cite{FlaCGEC} \\
        & & & & & EBGEC~\cite{Kaneko_Takase_Niwa_Okazaki}&BART~\cite{Shao_Geng_Liu_Dai} &GECToR~\cite{Omelianchuk_2020} \\
        \hdashline
        \multirow{2}{*}{Falko-MERLIN$\ast$} & \multirow{2}{*}{GE} & MERLIN, FalkoEssayL2, & \multirow{2}{*}{\cmark} & \multirow{2}{*}{$24$K} &$76.0$~\cite{Rothe_Mallinson_2021} &$74.5$~\cite{cao-etal-2023-unsupervised} &$73.7$~\cite{naplava-etal-2022-czech} \\ 
        & & FalkoEssayWhig & & &gT5~\cite{Rothe_Mallinson_2021} &BIFI~\cite{cao-etal-2023-unsupervised} &AG~\cite{naplava-etal-2022-czech} \\
        \hdashline
        \multirow{2}{*}{COWS-L2H$\ast$} & \multirow{2}{*}{ES} & \multirow{2}{*}{Essays} & \multirow{2}{*}{\cmark} & \multirow{2}{*}{$12$K} &$58.9$~\cite{stahlberg2024synthetic} &$57.3$~\cite{flachs2021data} &$55.2$~\cite{kementchedjhieva2023grammatical} \\
        & & & & &mT5~\cite{stahlberg2024synthetic} &Artificial~\cite{flachs2021data} &MT~\cite{kementchedjhieva2023grammatical} \\
        \hdashline
        \multirow{2}{*}{RONACC$\ast$} & \multirow{2}{*}{RO} & \multirow{2}{*}{TV and radio shows} & \multirow{2}{*}{\cmark} & \multirow{2}{*}{$10$K} &$69.0$~\cite{niculescu2021rogpt2} &$67.4$~\cite{kumar2024context} &$66.5$~\cite{kumar2024context} \\
        & & & & &RoGPT2~\cite{niculescu2021rogpt2} &AT-GCM~\cite{kumar2024context} &NAT-GEC~\cite{sun2022unified} \\
        \midrule 
        \multirow{2}{*}{Defects4J$\vartriangle$} & \multirow{2}{*}{Java} & \multirow{2}{*}{Open source programs} & \multirow{2}{*}{-} & \multirow{2}{*}{$357$} &$80/90$~\cite{yin2024thinkrepair} 
        &$79/50$~\cite{meng2023template} &$72/50$~\cite{jiang2023knod}\\
        & & & & &ThinkRepair~\cite{yin2024thinkrepair}  
        &TENURE~\cite{meng2023template}
        &KNOD~\cite{jiang2023knod} \\
        \hdashline
        \multirow{2}{*}{ManyBugs$\vartriangle$} & \multirow{2}{*}{C} & \multirow{2}{*}{Open source programs} & \multirow{2}{*}{-} & \multirow{2}{*}{$185$} &$16$~\cite{gharibi2024t5apr} &$15$~\cite{gharibi2024t5apr} &$12$~\cite{xia2023automated} \\
        & & & & &SOSRepair~\cite{afzal2019sosrepair} & T5APR~\cite{gharibi2024t5apr} &Codex~\cite{chen2021evaluating} \\
        \hdashline
        \multirow{2}{*}{IntroClass$\diamond$} & \multirow{2}{*}{C} & \multirow{2}{*}{Student programs} & \multirow{2}{*}{-} & \multirow{2}{*}{$998$} &$292/287$~\cite{koyuncu2020flexirepair} &$212/247$~\cite{koyuncu2020flexirepair} &$261/186$~\cite{koyuncu2020flexirepair} \\
        & & & & &GenProg~\cite{le2011genprog} &TrpAutoRepair~\cite{qi2013efficient} &FLEXIREPAIR~\cite{koyuncu2020flexirepair} \\
        \hdashline
        \multirow{2}{*}{QuixBugs$\vartriangle$} & \multirow{2}{*}{Java, Python} & \multirow{2}{*}{Quixey Challenge} & \multirow{2}{*}{-} & \multirow{2}{*}{$40$} &$39/40$~\cite{yin2024thinkrepair} 
        &$38/35$~\cite{yin2024thinkrepair}
        &$32/37$~\cite{yin2024thinkrepair}  \\
        & & & & &ThinkRepair~\cite{yin2024thinkrepair} 
        &BaseChatGPT~\cite{yin2024thinkrepair} &Codex~\cite{chen2021evaluating}  \\
        \hdashline
        \multirow{2}{*}{Bugs2Fix$\dagger$} & \multirow{2}{*}{Java} & \multirow{2}{*}{Github} & \multirow{2}{*}{-} & \multirow{2}{*}{$2.3$M} &$23.8$~\cite{gong2401ast} &$21.6$~\cite{gong2401ast} &$19.2$~\cite{gong2401ast} \\
        & & & & &AST-T5~\cite{gong2401ast} &CodeT5~\cite{wang2021codet5} &PLBART~\cite{Ahmad_Chakraborty_2021} \\
        \hdashline
        \multirow{2}{*}{CodeReview$\ddagger$} & Multi & \multirow{2}{*}{Github} & \multirow{2}{*}{-} & \multirow{2}{*}{$642$K} &$5.7$~\cite{lu2023llama} &$5.3$~\cite{lu2023llama} &$4.8$~\cite{lu2023llama} \\
        & (9 languages) & & & &LLaMA-Reviewer~\cite{lu2023llama} &CodeReviewer~\cite{CodeReview_2022} &CodeT5~\cite{wang2021codet5} \\
        \hdashline
        \multirow{2}{*}{CodeReview-New$\ddagger$} & Multi & CodeReview, & \multirow{2}{*}{-} & \multirow{2}{*}{$15$K} &$19.5$~\cite{guo2023exploring} &$14.8$~\cite{guo2023exploring} &\multirow{2}{*}{-} \\ 
        & (16 languages) & code reviews & & &ChatGPT &CodeReviewer~\cite{CodeReview_2022} & \\
        \bottomrule
    \end{tabular}
}
    \caption{Comparison of GEC and CEC datasets.
    ``Exp'' denotes whether the annotation contains the explicit error types.
    ``\#Q'' denotes the number of sentences in the dataset. 
    \changed{The evaluation metric is F$_{0.5}$$\ast$ score for GEC. For CEC, the metrics are Exact Match$\dagger$ for Bugs2Fix and CodeReview-New, Number of Correct Patches$\vartriangle$ for Defects4J, ManyBugs and QuixBugs, Number of Plausible Patches$\diamond$ for IntroClass, and BLEU-4$\ddagger$ for CodeReview.}
    }
\label{tab:datasetForLanguage}
\vspace{-0.5cm}
\end{table*}
\footnotetext{\url{https://gigaom.com/2012/01/19/quixey-challenge/}}

\begin{figure}[th!] 
	\centering
	\includegraphics[width=.5\textwidth]{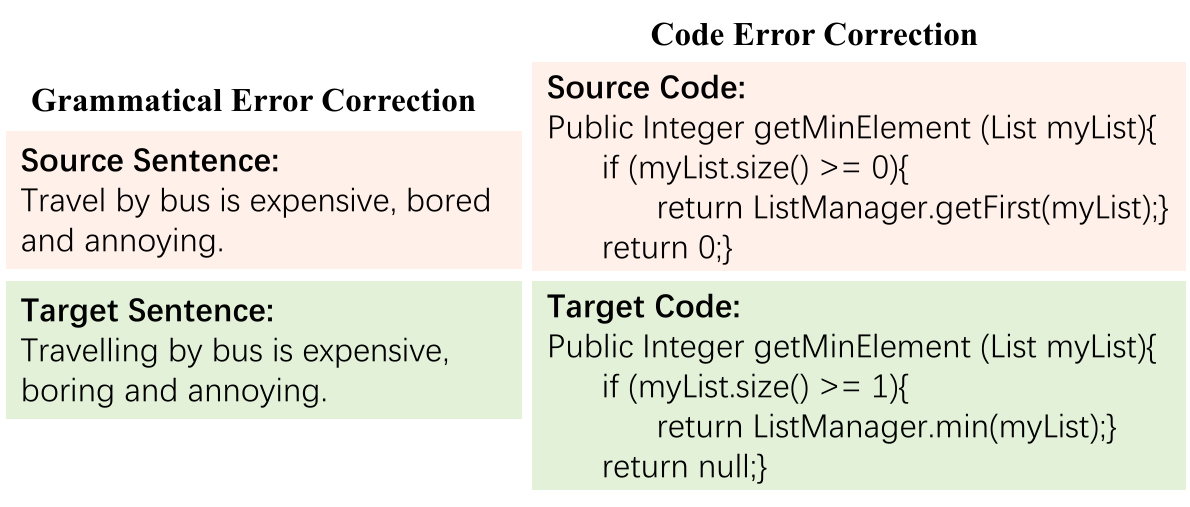} 
	\caption{The examples of GEC and CEC tasks. The examples are extracted from BEA-2019 and Bugs2Fix, respectively.}
	\label{fig:errorCorrection}
\end{figure}

Compared with AA, EC requires systems to provide explicit edits and corresponding revisions to the errors.
Similarly, language learning, as an important educational scenario, contains applications of Grammatical Error Correction (GEC) and Code Error Correction (CEC) for natural language learning and machine language learning.
In terms of the techniques of GEC and CEC tasks, the CEC differs from GEC due to their different syntax.
And we display two typical examples of them in Figure~\ref{fig:errorCorrection}.

\subsubsection{Grammatical Error Correction} 

\changed{GEC tasks are designed to identify and correct grammatical errors in text, including spelling errors caused by phonological confusion and visual confusion, as well as grammatical errors caused by incorrect use of grammatical rules.
\begin{align*}
L_{\text{GEC}} = -\mathbb{E}_{(X, Y) \in \mathcal{D}} \log P_{\theta}(Y \mid X)
\end{align*}
Here $X$ and $Y$ are the source sentence and target sentence, respectively.
For the GEC task, how to correctly identify and correct grammar errors is difficult, and how to avoid over-correction is challenging.
}

\noindent \textbf{Datasets.} 
Lang-8 is one of the most widely used datasets for language error correction, covering 80 languages, but its distribution of languages is very skewed, with Japanese and English being its most prevalent languages~\cite{ Rothe_Mallinson_2021}.
CLANG-8~\cite{Rothe_Mallinson_2021} is a clean dataset based on Lang-8, fixing some mismatches between data sources and targets in Lang-8. 
CoNLL-2014~\cite{Ng_Wu_Briscoe_2014} and BEA-2019~\cite{Bryant_Felice_2019} are both commonly used English language error correction datasets that refine different types of syntax errors.
In addition, with the popularity of GEC tasks, more and more non-English datasets have been proposed and widely used by researchers.
SIGHAN~\cite{Tseng_Lee_2015} and CCTC~\cite{wang2022cctc} are Chinese GEC datasets introduced to support more CGEC research.
The FCGEC~\cite{Xu_Wu_Peng_2022} and FlaCGEC~\cite{FlaCGEC} datasets have added more linguistic annotations, introduced more granular grammatical rules, and covered $28$ and $210$ error types, respectively, presenting new challenges to the GEC task.
At the same time, there are many GEC datasets for low-resource languages.
There are datasets GECCC~\cite{naplava-etal-2022-czech} in Czech, RULEC-GEC~\cite{Rozovskaya_Roth_2019} in Russian, Falko-Merlin~\cite{Roman_wiked2014} in German, COWS-L2H~\cite{Davidson_Yamada_Mira_2020} in Spanish, UA-GEC~\cite{Syvokon_Nahorna_2021} in Ukrainian and RONACC~\cite{Cotet_2020} in Romanian.

\noindent \textbf{Methods.} 
Early methods to handle GEC tasks are mostly achieved by manually defining rules or building classification-based models~\cite{Knight_Chander_1994, Rozovskaya_Chang_Sammons_2014}.
With the widespread use of neural networks, Seq2seq~\cite{Sutskever_Vinyals_2014} model has become the mainstream method to solve the task.
It treats grammatical error correction as a monolingual machine translation task, in which source sentences containing grammatical errors are tagged one by one to be translated into the correct target sentences.
The Seq2seq methods are mainly built based on RNN, CNN, or Transformer framework. 
BART~\cite{lewis2019bart}, T5~\cite{raffel2020exploring}, EBGEC~\cite{Kaneko_Takase_Niwa_Okazaki}, SynGEC~\cite{Zhang_Zhang_Li_2022} and CSynGEC~\cite{CSynGEC_2022} are good models in practice.
After that, Seq2edit~\cite{stahlberg-kumar-2020-seq2edits} methods have gradually emerged.
It treats the GEC task as a sequential labeling problem, marking the text spans with appropriate error tags, leaving the rest of the text unchanged, and generating one edit operation per prediction.
The commonly used efficient Seq2edit models include PIE~\cite{Awasthi_Sarawagi_2019}, EditNTS~\cite{dong-etal-2019}, LaserTagger~\cite{Malmi_Krause_2019}, GECToR~\cite{ Omelianchuk_2020} and so on.
Recently, the TemplateGEC~\cite{TemplateGEC_2023} model proposed by Li~\etal integrates the two frameworks of Seq2seq and Seq2edit, making use of their ability in error detection and correction, alleviating the problem of over-correction of seq2seq model to a certain extent, enhancing the effectiveness and robustness of the model. 
While GEC methods have achieved high performance, it still faces some problems and challenges.
\changed{
\begin{itemize}[leftmargin=*] 
\item \textbf{GEC with LLMs encounters over-correction issue.}
Even though LLMs are able to solve GEC tasks in nature, it has an over-correction issue, which hinders its application in the education domain since teachers are meant to correct the sentences with minimum edits.
Some studies change the role of LLMs in GEC, utilizing them to improve the performance of the smaller model~\cite{li2024rethinking}. Also, some studies focus on addressing the over-correction issue in LLMs directly. Wang~\etal~\cite{wang2024lm} used cross-inference to generate a set of correction candidates and specifically train a model for soft ensemble within the candidate set to avoid over-correction. Omelianchuk~\etal~\cite{omelianchuk2024pillars} used multiple models for voting to prevent over-correction in the output of a single model. 
\begin{align*}
\hat{Y} = \text{Ensemble}_{Y_i \sim P_{\theta}(Y_i | X)}(Y_i).
\end{align*}
\item \textbf{Distorted results of GEC systems in low resources.}
For language teaching over the world, the development of GEC systems for low-resource languages is underdeveloped.
Many studies focus on applying current GEC methods to low-resource languages, such as Bangla~\cite{BanglaGEC}, Slovak~\cite{slovakGEC}, Turkish~\cite{2024gecturk}, Zarma~\cite{keita2024grammatical} and so on. Chan~\etal~\cite{chan2024grammatical} proposed a method for constructing multilingual data to handle mixed-language text, which significantly improves the performance of GEC systems.
\end{itemize}
\ysadd{Previous studies have not considered the GEC with visual features, where hand-written essays are submitted by students to be corrected.
Besides, the evaluation criterion of GEC following minimum edits should be reconsidered in real-world since different corrections maybe acceptable.}
}

\subsubsection{Code Error Correction} 


\changed{
Code Error Correction (CEC) fixes a buggy code snippet to make it error-free and more coherent. 
It is a challenging task as it requires the model to have a sufficient understanding of the code in long-distance dependency and make proper corrections to the erroneous code spans.
It shares similar principles as GEC tasks and we can formulate both of them as neural machine translation~\cite{han:sigir2023,chen:emnlpfinding2023}:
\begin{align*}
L_{\text{CEC}} = -\mathbb{E}_{(X, Y) \in \mathcal{D}} \log P_{\theta}( Y \mid X)
\end{align*}
Here, $X$ and $Y$ are the input and output code sequences.
}


\noindent \textbf{Datasets.} 
Defects4J~\cite{Just_Jalali_Ernst_2014}, ManyBugs~\cite{LeGoues_Holtschulte_2015} and IntroClass~\cite{LeGoues_Holtschulte_2015} are some of the earlier datasets proposed for CEC tasks, focusing on Java and C languages respectively, but with small amounts of data.
Lin~\etal~\cite{Lin_Koppel_Chen_2017} proposed the first multi-lingual parallel corpus of CEC benchmarks, namely QuixBugs\cite{Lin_Koppel_Chen_2017}, which contains $40$ buggy code snippets translated into both Java and Python.
Xia~\etal~\cite{Xia_Wei_Zhang_2022} combined the aforementioned Defects4J, ManyBugs, and QuixBugs datasets into a single CEC evaluation dataset to test the effect of PLM on the CEC task.
Bugs2Fix~\cite{Tufano_Watson_2018} is a dataset mined from GitHub bug-fixing commits, containing 2.3M bug-fix pairs (BFPS).
CodeReview~\cite{CodeReview_2022} and CodeReview-New~\cite{guo2023exploring} are also large-scale datasets of code reviews collected from open-source projects, involving nine of the most popular programming languages.

\noindent \textbf{Methods.}  
Existing techniques capture the structural and semantic information of code by representing it as a sequence of symbols or an Abstract Syntax Tree (AST)~\cite{AST_2014}, which involves some unique challenges for CEC tasks.
\begin{itemize}[leftmargin=*]
\item \textbf{The structural information of code has been ignored.}
Both CuBERT~\cite{kanade:pmlr2020} and CodeBERT~\cite{Feng_Guo_Tang_2020} ignore the structural information of code. 
In response to this issue, Guo~\etal~\cite{Guo_Ren_Lu_Feng_2020} proposed the GraphCodeBert model to represent source code information based on AST data flow, so as to reflect deeper code information.
GraphCodeBERT adapts Transformer and allows the model to learn the AST data flow.
The follow-up UniXcoder~\cite{Guo_Lu_Duan_Wang} improves GraphCodeBERT by converting an AST into a sequence structure.
PLBART~\cite{Ahmad_Chakraborty_2021} incorporates the denoising strategies to better understand program grammar and logical flows.
CodeT5~\cite{wang2021codet5} uses the T5 model architecture and makes use of the identifier tagging and comment information to better model the structural characteristics of code. 

\item \textbf{CEC systems have resource overhead issue.}
CEC systems' high demands for memory and computing power may disrupt the user's workflow and decrease productivity~\cite{anand2024comprehensivesurveyaidrivenadvancements}. CodePAD~\cite{Dong_Jiang_Liu_2022} designs an approach based on pushdown automaton (PDA), which takes into account the grammar constraints of programming languages and reduces the unnecessary computational cost.

\item \textbf{CEC systems encounter overfitting issue}
Overfitting refers to the issue where generated patches only pass the developer-written test suite but fail to generalize to other potential test cases~\cite{zhang2023surveylearningbasedautomatedprogram}. 
Nilizadeh~\etal~\cite{nilizadeh2021exploring} introduced a method that identifies correct patches based on the similarity of test case execution between the buggy and patched programs, which reduces overfitting by evaluating patches across a broader range of execution scenarios. Ye~\etal~\cite{Ye_2022} proposed an overfitting detection system, which leverages static code features, such as variable types, code structure, and contextual syntax information, to classify patches as overfitting or correct. 
\end{itemize}
\ysadd{The performance and memory load of CEC systems could be improved more with the consideration of the informal and rigorous teaching environment for coding teaching.}


\section{Demonstration Systems}
\label{sec:demo}

\begin{table*}[t!]
\centering
\small
\begin{tabular}{l l l l l l l l}
    \toprule  
    Task & Demos &  & & & & & \\
    \midrule
    QA  & \href{https://docs.openvino.ai/2023.3/notebooks_section_2_model_demos.html}{OpenVINO~$\dagger$} 
         & \href{https://github.com/primeqa}{PrimeQA~$\dagger$} 
         &\href{https://github.com/lfy79001/TableQAKit}{TableQAKit~$\dagger$} 
         & \href{https://github.com/arunmallya/piggyback}{PiggyBack~$\ast$} 
         & \href{https://ascent.mpi-inf.mpg.de/}{ASCENT~$\diamond$}
          & \href{https://square.ukp-lab.de/}{UKP-SQUARE~$\diamond$}
          & \href{https://github.com/jasonyux/LocalRQA}{LOCALRQA~$\ast$}\\
    \hdashline
    MWP & \href{https://github.com/LYH-YF/MWPToolkit}{MWPToolkit~$\dagger$} 
         & \href{https://www.intmath.com/}{IntMath~$\diamond$} 
         &  \href{https://github.com/goelm08/MWP-ranker}{MWPRanker~$\ast$}& \\
    \hdashline
    QC  & \href{https://github.com/asahi417/lm-question-generation}{AutoQG~$\ast$} 
         & \href{https://github.com/StevenJamesMoore/AIED24}{SAQUET~$\dagger$} & \href{https://github.com/roemmele/answerquest}{AnswerQuest~$\diamond$} & \href{https://github.com/KristiyanVachev/Leaf-Question-Generation}{Leaf~$\ast$} \\
    \hdashline
    AA & \href{https://f.linggle.com/}{LinggleWrite~$\ast$} 
         & \href{https://www.intellimetric.com/direct/}{IntelliMetric~$\diamond$} & \href{https://github.com/octanove/expats}{EXPATS~$\dagger$}  & \href{https://www.qualified.io/}{Qualified~$\diamond$} 
         & \href{https://coderpad.io/}{CoderPad~$\diamond$} & & \\
    \hdashline
    GEC & \href{https://sterling8.d2.comp.nus.edu.sg/allecs/}{ALLECS~$\diamond$} 
         & \href{https://www.grammarly.com/}{Grammarly~$\diamond$} 
         & \href{https://github.com/chrisjbryant/errant}{ERRANT~$\dagger$}
         & \href{https://github.com/nusnlp/wamp}{WAMP~$\ast$}
         &\href{https://github.com/thiborose/gecko-app}{GECko+~$\ast$} 
         & \href{https://effidit.qq.com/}{Effidit~$\diamond$}\\
    \hdashline
    CEC & \href{https://codegeex.cn/}{CodeGeeX~$\vartriangle$} 
         & \href{https://codehelp.app/}{CodeHelp~$\diamond$} & \href{https://iflycode.xfyun.cn/index}{iFlyCode~$\diamond$} &\href{https://www.marscode.com/}{MarsCode~$\vartriangle$}\\
    \bottomrule
\end{tabular}
\caption{Demos that can be applied in education domain. \changed{$\diamond$: platform, $\ast$: application, $\dagger$: toolkit, $\vartriangle$: extension. We regard a platform as a tool that can be directly used by educators, while an application is a tool that requires additional setup and deployment before it can be utilized. A toolkit is an extensible collection of tools, functions, and resources designed for specific tasks. An extension is a tool that exists in the form of a plugin.}
}
\label{tab:demo}
\vspace{-0.5cm}
\end{table*}

\changed{To facilitate applications on education scenarios, in this section, we present some off-the-shelf demo cases with brief illustrations of their usage for each task mentioned in Section 3. 
Information on these demos will be presented in Table~\ref{tab:demo}.
We annotate demonstrations with ``platform'', ``application'', ``toolkit'' and ``extension''.
Teachers and students could utilize platforms and extensions featured with interfaces and user instructions as AI helpers.
Researchers could utilize toolkits and applications featured with development frameworks and flexible modules to implement new methods quickly.
}

\changed{For QA, OpenVINO\footnote{\url{https://docs.openvino.ai/2023.3/notebooks_section_2_model_demos.html}} is an open-source toolkit designed for diverse automatic QA, including table, visual, interactive, and LLMs-based QA. PrimeQA~\cite{sil2023primeqa} is an open-source public repository designed to democratize QA research and simplify the replication of state-of-the-art methods, supporting the training of multilingual QA models with core functionalities like information retrieval, reading comprehension, and question generation. TableQAKit~\cite{lei2023tableqakit} is an open-source toolkit designed for table-based QA, offering a unified platform with various datasets, methods, and LLM integrations. PiggyBack~\cite{zhang2023piggyback} is a user-friendly VQA platform that simplifies the use of visual-language pre-trained models with a browser-based interface and comprehensive task support. 
Besides, other QA demos~\cite{nguyen2021inside, baumgartner2022ukp, yu2024localrqa} are also available for both educators and researchers.
}

\changed{For MWP, MWPToolkit~\cite{lan2022mwptoolkit} offers a holistic and extensible framework that provides deep learning-based solvers, popular datasets, and a modular architecture that supports quick replication and innovation of MWP methods. IntMath\footnote{\url{https://www.intmath.com/}} is an online math platform that combines a mathematical computation engine with artificial intelligence to parse and generates step-by-step natural language answers. MWPRanker~\cite{goel2023mwpranker} is a tool to retrieve similar math word problems based on expression tree similarity.
}

\changed{For QC, AutoQG~\cite{ushio2023practical} is a service that provides multilingual question-and-answer generation, featuring a comprehensive toolkit for model fine-tuning, generation, and evaluation. SAQUET~\cite{moore2024automaticquestionusabilityevaluation} proposes a toolkit to assess the structural and pedagogical quality of MCQs. AnswerQuest~\cite{roemmele2021answerquest} is a system that integrates QA and QG to create Q\&A items for enhancing reading comprehension of multi-paragraph documents. Besides, other QC demos~\cite{vachev2022leaf} are also available for both educators and researchers.
}

\changed{For AES and ACS, LinggleWrite~\cite{tsai2020lingglewrite} is a writing coach tool that offers writing suggestions, proficiency assessments, grammatical error detection, and corrective feedback. IntelliMetric\footnote{\url{https://www.intellimetric.com/direct/}} is an online multilingual AES platform that supports writing evaluation by delivering instant scores with the accuracy and consistency of human experts. EXPATS~\cite{manabe2021expats} is an open-source framework for AES, enabling rapid experimentation with various models, while integrating visualization and interpretability tools. Qualified\footnote{\url{https://www.qualified.io/}} and CoderPad\footnote{\url{https://coderpad.io/}} are online platforms designed for coding assessments, offering interactive environments to evaluate coding skills.
}

For GEC and CEC, ALLECS~\cite{qorib2023allecs} and Grammarly\footnote{\url{https://www.grammarly.com/}} are web applications that provide online GEC service by offering lightweight GEC systems optimized for users with slow internet connections and comprehensive writing assistance, respectively. Errant~\cite{bryant2017automatic} is a toolkit that extracts and classifies edits from parallel sentences, enabling error type evaluation and dataset standardization. WAMP~\cite{moon2023wamp} is a web-based annotation tool designed to efficiently create annotated corpora for GEC, with features like customizable error tags and file export for system evaluation. Other demos~\cite{calo2021gecko+, shi2023effidit} can also be applied for both researchers and educators. CodeGeeX~\cite{zheng2023codegeex} is an open-source and multilingual pre-trained model that enables syntax and function-correct code generation and can be integrated as IDE extensions. CodeHelp~\cite{Sheese_2024} is an LLM-powered assistant that provides real-time programming guidance and explanations, helping students solve coding issues independently without giving direct solutions. MarsCode~\cite{liu2024marscodeagentainativeautomated} is an LLM-based multi-agent framework for automated bug fixing.

\section{Challenges and Future Trends}



In this section, we will discuss the challenges and future trends when applying NLP technology in real educational scenarios.

\noindent \textbf{Generalization over subjects and languages}.
Even though there is a range of datasets from question answering to error correction, we also observe there is still a demand for datasets and techniques fitting in more subjects and languages.
For example, for AES and GEC tasks, there is a lack of AES and GEC datasets in non-English languages, which is vital for language education in non-English countries.
For CEC tasks, we found limited datasets for Python language, which is a widely used programming language today.
Similarly, applications on different subjects are vital in extending the NLP techniques to education.
Even though there is a series of datasets covering the subjects of grades 6-8 or K-12, for higher education, which includes some advanced subjects like philosophy, and history, we still face a shortage of data.
In the future, a comprehensive dataset is desired to cover subjects of different learning stages and different tasks for both teaching and research.

\noindent \textbf{Deployed LLM-based systems for education}.
Even though an individual task can be well-formulated and solved via specific techniques, there is a demand for integrated systems for education.
Baladn~\etal~\cite{baladn:bea2023} tuned open-source LLMs for generating teacher responses in BEA 2023 Shared Task.
EduChat~\cite{dan:arxiv2023} is a more general LLM in the education domain, which is able to automatically assess essays, provide emotional support, and conduct Socratic teaching with the forms of chatting and question answering.
In the future, an integrated LLM-based tutoring system should be featured with stronger capabilities like answering textbook questions, and automated quiz-making in the dimension of different subjects such as math, computer science, linguistics, and so on.
\changed{Regarding system deployment, 
although open-source NLP technology demos, including toolkits, online platforms, and applications, have been widely developed (see Section~\ref{sec:demo}).
The limited hardware and resources make it challenging to apply advanced NLP in education, prompting studies focused on improving feasibility in this area. 
For example, studies~\cite{fan2022framework, lin2024large} of MWP solving have shown that applying knowledge distillation can significantly reduce model size and improve inference speed while maintaining high solution accuracy. 
Further optimization is needed considering the edge computing and efficiency.
}

\noindent \textbf{Adaptive learning for teaching and learning}.
Even though NLP applications can be applied to the education domain, there are a few studies involving adaptive learning.
In the QG task, we notice some researchers~\cite{srivastava2021question,wang2023difficulty} attempt to model the learning trajectories of students by incorporating their historical answers to predict their performance on the new questions, which are deemed as a significant feature to determine the difficulty level of the next question.
This helps to build a more customized education system.
Such adaptive learning is in high demand to collaborate with other tasks.
For example, in EC tasks, when we correct the sentences and codes, we can consider the learning trajectory of the students to decide the way to correct them.
If it is a simple sentence with a misused word, we should not substitute it with an obscure word.
Besides modeling the learning trajectories, the intervention of difficulty-level control mechanisms should be integrated into the education system.
Even though there are multiple studies~\cite{gao2018difficulty,cheng2021guiding,uto2023difficulty} trying to model the difficulty level in the QG task, they fail to align the level to the difficulty level of the syllabus.

\noindent \textbf{Interpretability for education}.
Existing studies are able to obtain good results on various tasks like question answering, question construction, automated assessment, and error correction.
However, there are limited techniques developed with the perspective of interpretability.
As shown in the technical review of QA tasks, recent studies try to explicate the process of thinking in solving TQA and MWPs tasks~\cite{andreas:arxiv2016,Wei:arxiv2023}.
In other tasks, interpretability should also be taken into account.
It would be promising to show a thinking path and make the construction of the questions more rational.
For AS and EC tasks, showing a fine-grained explanation of the scoring and correction would not only improve the tutoring experience but also benefit the methods as well as systems to diagnose their functions.
Even though we notice a few studies have started to develop interpretable GEC systems~\cite{FlaCGEC}, interpretability should be further studied and explored with more attention.

\noindent \textbf{Ethical considerations of NLP techniques}.
\changed{Applying NLP techniques for education raises various ethical issues, such as fairness, safety, privacy, and dependency. 
Li~\etal~\cite{li2024surveyfairnesslargelanguage} systematically introduce the fairness of LMs. 
As noted by Chinta et al.~\cite{chinta2024fairaied}, racial bias in AI-driven essay scoring systems has resulted in darker-skinned students receiving lower scores than other students for essays of comparable quality.
Regarding privacy concerns, when students use third-party generative AI tools for assignments or reflections, they may unintentionally expose sensitive academic data, risking privacy law violations. These tools lack transparency around how data is stored, processed, or reused~\cite{ng2025analyzingsecurityprivacychallenges}. To address this issue, researchers~\cite{latif2025privacypreservedautomatedscoringusing, khalil2025towards} have shown that federated learning protects student privacy by training models locally on sensitive data and sharing only model updates, enabling collaborative model development without exposing raw data.
For dependency, Krupp~\etal~\cite{krupp2023challengesopportunitiesmoderatingusage} indicate an overreliance on ChatGPT among prospective physics teachers and students, affecting their use of real-world contextualization and leading to poorer results in complex tasks. Multiple education-related studies also attempt to reduce educators’ or students’ overreliance on current techniques~\cite{Sheese_2024, zhai2024effects}. 
}

\section{Conclusions}

This survey attempted to provide a comprehensive overview of NLP in the education domain.
We highlight that NLP techniques can be applied to various procedures of education.
Our survey highlights four representative tasks (i.e., question answering, question construction, automated assessment, and error correction) with their fine-grained sub-tasks as the focus of this survey due to their significant impact on teaching and learning.
According to the taxonomy, we go deep into these tasks and illustrate the evolution of their techniques.
Despite the strides made in NLP applications in the education domain, there still remains room for the improvement of diverse tasks.
In the last section, we suggest some future directions for NLP in the education domain.

We would like to note that besides the highlighted tasks in this survey, there are other studies that can contribute to the educational scenarios.
For example, knowledge graph construction~\cite{wright:2022bioact} aims to organize and extract the knowledge points from the unstructured data sources to form a knowledge graph, which is able to demonstrate the connection between the schema for various subjects and support downstream applications.
In summary, the demand for educational applications is increasing and educational NLP will thrive to be a promising research area.
We hope this survey will give a comprehensive picture of educational NLP and we encourage more contribution in this field.

\bibliographystyle{IEEEtran}
\bibliography{sn-bibliography}


%

\ignore{
\appendices
\section{Proof of the First Zonklar Equation}
Appendix one text goes here.

\section{}
Appendix two text goes here.
}

\ignore{
\ifCLASSOPTIONcompsoc
  \section*{Acknowledgments}
\else
  \section*{Acknowledgment}
\fi

The authors would like to thank...
}

\ifCLASSOPTIONcaptionsoff
  \newpage
\fi

\end{document}